\documentclass{article}

\usepackage{iclr2027_conference, times}

\usepackage[utf8]{inputenc} % allow utf-8 input
\usepackage[T1]{fontenc}    % use 8-bit T1 fonts
\usepackage{hyperref}       % hyperlinks
\usepackage{url}            % simple URL typesetting
\usepackage{booktabs}       % professional-quality tables
\usepackage{amsfonts}       % blackboard math symbols
\usepackage{nicefrac}       % compact symbols for 1/2, etc.
\usepackage{microtype}      % microtypography
\usepackage{xcolor}         % colors
\usepackage{multirow}   
\usepackage{amsmath}
\usepackage{enumitem}
\usepackage{graphicx}
\usepackage{algorithm}
\usepackage{algpseudocode}
\usepackage{wrapfig}
\usepackage{float}
\usepackage{longtable}
\usepackage{fvextra}

\graphicspath{ {./figures/} }

\title{SAVE: Less is More in Agentic Video Reasoning}

\author{Tymoteusz Wilk \\
Technical University of Munich \\
\texttt{tymoteusz.wilk@tum.de} \\
\And
Sergey Prokudin\textsuperscript{*} \\
ETH Zurich \\
\texttt{sergey.prokudin@inf.ethz.ch} \\
\AND
Xi Wang\textsuperscript{*} \\
ETH Zurich, Technical University of Munich, \\ MCML, Ulm University \\
\texttt{xi.wang@inf.ethz.ch}
}

\iclrfinalcopy

\begin{document}

\maketitle
\lhead{}
\begingroup
\renewcommand{\thefootnote}{\fnsymbol{footnote}}
\footnotetext[1]{Equal advising}
\endgroup

\begin{abstract}
    Agentic video-reasoning frameworks iteratively explore a video, choosing what to observe based on the query and evidence gathered so far. Existing methods are typically evaluated with different backbones, preprocessing pipelines, and computational budgets, making it difficult to determine which components actually drive performance, how much they cost, and whether agentic inference is preferable to native video reasoning in the first place. We conduct the broadest controlled evaluation to date of a diverse set of recent agentic video-reasoning systems across two strong contemporary backbones, with explicit token and monetary-cost accounting. Across Gemini~3 Flash and Gemma~4, most existing agentic frameworks \emph{fail} to consistently improve over native video inference and often increase computational cost. We investigate two primary questions: (1) How much agentic framework complexity is necessary for effective video reasoning? (2) When, if at all, is agentic video reasoning preferable to just using the backbone directly? To answer the first question, we introduce a deliberately minimal Active Exploration Baseline and show that it remains competitive with substantially more elaborate agents. We then propose targeted extensions to this baseline, yielding SAVE (Simple Agentic Video Explorer). SAVE outperforms all evaluated prior agentic methods in five of six backbone--benchmark settings. On Minerva, SAVE reaches 66.4\% accuracy with Gemini~3 Flash, comparable to 64.8\% for native inference, while reducing inference cost by 60\%, and exceeding the highest-accuracy evaluated prior agentic baseline by 9.9 percentage points with 79\% lower cost. Motivated by this fact, we ask when agents are preferable to simply using the underlying backbone. Our results indicate that the advantage of agentic exploration grows when dense processing would otherwise represent answer-relevant events too sparsely, particularly for long videos and temporally concentrated evidence. Overall, our results suggest that agentic video reasoning should favor simple frameworks over increasingly elaborate pipelines and focus primarily on long-form video, where selective evidence acquisition offers the clearest advantage over native inference.
\end{abstract}

\section{Introduction}

Long-form video reasoning remains a fundamental challenge for modern vision-language models because it requires processing large amounts of temporally structured information, performing complex visual reasoning, and integrating evidence over long time horizons. In videos, the observations needed to answer a question may be distributed across thousands of frames, creating a tension between reasoning quality and computational efficiency: densely processing the entire video is expensive, while aggressive downsampling or compression risks discarding critical evidence. Recent multimodal large language models (MLLMs), including GPT-4o~\citep{gpt4o}, and Gemini~3~\citep{gemini3_2026}, have substantially improved video understanding, but efficient and accurate reasoning remains challenging.

A growing body of work addresses this problem through agentic video reasoning, where models iteratively select observations conditioned on the query and previously gathered evidence~\citep{zuo2025videolucy,pang2025mrvideo,wang2025videotree,zhi2025videoagent2}. While previous work asks the question “How do we build better video agents?”, we ask two complementary questions: (1) How much agentic machinery is actually necessary? and (2) When, if at all, is agentic inference preferable to simply using the underlying video backbone directly? Existing methods are typically evaluated with different multimodal backbones, computational budgets, preprocessing pipelines, and benchmark settings. This makes it difficult to determine which components of the agentic framework actually drive performance, what constitutes an effective video agent, and whether agentic inference is preferable to native video reasoning in the first place.

To address those questions, we conduct, to the best of our knowledge, the most comprehensive controlled evaluation to date of contemporary agentic video reasoning frameworks under shared frontier backbones. We systematically evaluate both Gemini~3~Flash~\citep{gemini3_2026} and the open-weight Gemma~4~\citep{gemma4}, adapting a diverse set of recent methods to use the same underlying backbone, evaluating them on a common suite of video reasoning benchmarks, and explicitly accounting for computational cost. Surprisingly, we find that when paired with strong multimodal backbones, most existing agentic frameworks \emph{fail} to consistently outperform native video inference and frequently incur substantially greater computational cost. We observe this trend across multiple datasets and agentic formulations.

Despite their differences, most video agents share a common structure: they acquire partial observations, update a reasoning state, and decide what to inspect next. We capture this structure with a simple abstraction inspired by partially observable Markov decision processes (POMDPs). To answer the first question, we instantiate it with a deliberately minimal baseline in which the backbone itself selects individual frames to inspect, without specialized retrieval, hierarchical memory, or video-wide preprocessing. Despite its simplicity, this baseline remains competitive with substantially more elaborate systems.

Building on this observation, we introduce \textit{SAVE (Simple Agentic Video Explorer)}, which improves over the minimal baseline using targeted adjustments.  SAVE preserves the same simple interaction loop, but lets the backbone query temporal ranges and adaptively allocates temporal and visual resolution within them. Across three benchmarks and two backbones, SAVE achieves the highest performance among evaluated prior agentic methods in five of six settings, while generally using substantially fewer resources. On Minerva, for example, SAVE reaches 66.4\% with Gemini~3~Flash, comparable to 64.8\% for native inference and significantly higher than 56.5\% for VideoLucy, while reducing monetary inference cost by approximately 60\% and 79\%, respectively.

Finally, we ask when agentic exploration helps. On Minerva, SAVE's advantage over a token-matched native baseline grows with video duration. A controlled synthetic experiment further shows that adaptive inspection is most useful when the temporal pattern needed to answer a question occupies only a small fraction of the video. Together, our results suggest a simple principle: agentic exploration is most useful when uniform processing would represent answer-relevant evidence too sparsely. Agentic video reasoning appears to be most promising in the long-form regime, where selective evidence acquisition provides the clearest advantage over dense native processing; for shorter videos, the additional overhead of agentic exploration may offer limited benefit.

To summarize, our contributions are threefold: (1) We perform, to the best of our knowledge, the most comprehensive controlled evaluation to date of agentic video reasoning methods under shared frontier backbones with explicit computational accounting, showing that most evaluated frameworks fail to improve over native inference while often substantially increasing cost. (2) We introduce a deliberately minimal Active Exploration Baseline, and build on this baseline with SAVE. Despite its simplicity, the baseline remains competitive with substantially more elaborate agents, while SAVE outperforms all prior agentic approaches in most settings at substantially lower computational cost. (3) We characterize the regime in which agentic exploration is most useful, namely when uniform processing would represent answer-relevant evidence too sparsely.

\section{Related Work}

\paragraph{MLLMs for video reasoning.}
Early video MLLMs such as Video-ChatGPT~\citep{Maaz2023VideoChatGPT},
VideoChat~\citep{2023videochat}, and Video-LLaVA~\citep{lin2023video, zhu2023languagebind}
established the ability of language models to reason over sequential visual inputs.
More recent systems, including proprietary models such as Gemini~3~\citep{gemini3_2026},
open-weight models such as Gemma~3~\citep{gemmateam2025gemma3technicalreport}, and other recent video MLLMs~\citep{wang2025internvideo,
damonlpsg2023videollama, damonlpsg2024videollama2,
damonlpsg2025videollama3, bai2025qwen3vltechnicalreport},
have substantially improved video understanding and reasoning capabilities. Despite these advances, video reasoning remains constrained by a fundamental trade-off between computational cost and information retention. Processing dense visual inputs preserves fine-grained evidence but incurs substantial memory and compute costs, while uniform downsampling, token compression~\citep{bolya2023tome}, or pruning~\citep{chen2024fastv} reduce context size at the risk of discarding question-relevant information. This limitation motivates explicit, query-dependent frame-selection methods.

\paragraph{Frame selection.}
Frame-selection methods typically operate on top of a vision-capable MLLM backbone and differ primarily in how directly that backbone controls evidence acquisition. We call methods \emph{agentic} when the backbone adapts the high-level search policy based on intermediate observations. Non-agentic methods instead follow a predefined selection procedure before the final reasoning step. We refer to the former as backbone-driven \emph{active exploration} and note that this distinction is a continuum rather than a strict dichotomy.

\textit{Non-agentic approaches} include T*~\citep{ye2025trethinkingtemporalsearch}, AKS~\citep{tang2025aks}, and FOCUS~\citep{zhu2025focus}. Their main advantage is that the expensive final backbone typically processes the selected evidence only once, while frame selection is delegated to predefined and often complex pipelines built around lighter auxiliary models such as CLIP~\citep{radford2021learning}. While this makes keyframe selection efficient, it also limits the selector to the representations and heuristics encoded by these auxiliary components. As a remedy, recent approaches such as TCoT~\citep{arnab2026temporal} use the VLM itself in the frame selection pipeline at the cost of using more computational resources.

\textit{Agentic approaches} have been gaining popularity as an alternative to these fixed selection pipelines. Some first employ an initial, question-agnostic preprocessing step, e.g., to construct a searchable index or subject registry of the entire video. VideoLucy \citep{zuo2025videolucy}, DeepVideoDiscovery \citep{zhang2025deep}, VideoAgent2 \citep{zhi2025videoagent2} and MR Video \citep{pang2025mrvideo} follow this paradigm. This step effectively narrows the temporal search space, providing significant benefits for long-form videos that require solving ``needle in a haystack'' localization problems. However preprocessing is computationally expensive, and because it's performed independently of the question, it risks omitting critical fine-grained details, potentially misguiding the agent's subsequent search. Others rely more on direct query-conditioned exploration. Building on early programmatic precursors \citep{surisvipergpt}, methods such as VideoAgent \citep{VideoAgent}, VideoCuriousAgent \citep{yang2025vca}, VideoTree \citep{wang2025videotree}, VideoARM \citep{yin2026videoarmagenticreasoninghierarchical}, and others \citep{zou2026air,lvagentchen2025,wang2025activevideoperceptioniterative}, perform actions on the raw video conditioned simultaneously on the query and the previously gathered evidence. VideoZoomer~\citep{ding2026videozoomer} is conceptually closely related to our work, as it likewise performs iterative coarse-to-fine temporal exploration; however, it learns a dedicated exploration policy through supervised fine-tuning and reinforcement learning.

% Complexity point

Across frame-selection paradigms, we observe a trend in the recent literature toward increasing complexity within the framework itself. Compared with the relatively simple interaction loop of VideoAgent~\citep{VideoAgent}, DeepVideoDiscovery~\citep{zhang2025deep} first constructs a multi-granular video database and exposes several specialized tools, including global browsing, clip-level search, and frame inspection, which the agent dynamically composes during inference. VideoLucy~\citep{zuo2025videolucy} goes further by maintaining hierarchical coarse-to-fine memories and coordinating multiple specialized agents for localization, instruction generation, captioning, and answering within an iterative backtracking procedure.

Prior work has primarily asked how to build better video agents. However, existing frameworks differ substantially in their formulation, inference budgets, preprocessing pipelines, and underlying backbones, making it difficult to isolate where their gains come from. As a result, two complementary questions remain largely unresolved: (1) How much additional machinery is actually necessary for effective video reasoning? and (2) When does agentic inference provide benefits over using the underlying backbone directly? The focus of our work is to disentangle these factors and answer both questions in a controlled setting.

\section{Active Exploration Framework}

In this section, we unify various agentic formulations under a single framework. Building upon this foundation, we first establish an Active Exploration Baseline as a minimal instantiation of this framework. We then introduce our proposed approach, SAVE, which addresses the weaknesses of the minimal baseline.

\subsection{Problem Formalization}

We observe that most existing agentic approaches can be viewed as instances of a unified framework. To identify the common structure behind those approaches, we 
present a mathematical formalization, drawing inspiration from partially observable Markov decision processes (POMDPs) \citep{KAELBLING199899}. %Our formalism can be seen as a generalization of the idea proposed in ReAct \citep{yao2023react}.
To demonstrate its generality, this formulation is applied to popular contemporary methods in Appendix~\ref{sec:formalism_application}.

Specifically, task-conditioned video inference is modeled as \emph{active exploration}.
A problem instance is modeled as \(E, q, y\)
where \(E\in\mathcal{E}\) is the environment (e.g., a video), \(q\in\mathcal{Q}\) is the user query, and \(y\in\mathcal{Y}\) is the ground-truth answer. In the video reasoning case, \(E=V=\{(x_i,t_i)\}_{i=1}^N\), where \(x_i\) denotes the \(i\)-th frame and \(t_i\) its timestamp.

\textbf{Actions and Observations.}
We define an observation space \(\mathcal{O}\) (e.g., frames, crops, text), an action space
\(
\mathcal{A}=\mathcal{A}_{\text{query}}\cup\{\textsc{Stop}\},
\)
and an intermediate output (state) space \(\mathcal{S}\). At each step the system can observe the environment through a deterministic operator:
\(
O:\mathcal{A}_{\text{query}}\times \mathcal{E}\rightarrow \mathcal{O}.
\)

\textbf{Internal state.}
We call \(s_t\) the agent state and \(a_t\) the action planned at time \(t\).
At timestep \(t+1\), given the agent state for time \(t\) and \(a_t\), the system executes the action \(o_{t+1} = O(a_t, E)\), and computes:
\(s_{t+1} = F(q, s_t, o_{t+1}) \in \mathcal{S}\). The state update \(F\) is defined as:
\(
F:\mathcal{Q} \times \mathcal{S} \times \mathcal{O} \rightarrow\mathcal{S}
\), \(F\) aggregates the query \(q\) and the evidence collected so far (captured by \(s_t\)) together with the newly acquired observation \(o_{t+1}\) into an intermediate representation \(s_{t+1}\) that is used both for action selection and subsequent predictions. No assumptions are made about the internal state in terms of size and scope; \(s_{t+1}\) may also include the entire history of interactions.

\textbf{Initialization.}
We define time as starting at \(t=1\), before any query action has been executed. The initial
state is produced by
\(
s_1 = F_{\text{init}}(q,E).
\)
This general form permits both query-dependent initialization and question-agnostic video
preprocessing; methods without video-wide preprocessing may simply ignore \(E\).

\textbf{Policy and termination.}
The next action, \(a_{t+1}\in\mathcal{A}\), is sampled from a policy
\(
a_{t+1} \sim \pi(\cdot \mid s_{t+1}).
\)
If \(a_{t+1}\in\mathcal{A}_{\text{query}}\), the process continues with \(a_{t+1}\) along with \(s_{t+1}\) being passed to the next iteration.
If \(a_{t+1}=\textsc{Stop}\), the interaction terminates at time \(t+1\). Upon termination, the system outputs a final answer using a prediction head:
\(
\hat{y} = g(s_{t+1}) \in \mathcal{Y}.
\)

% \paragraph{Objective under acquisition cost.}
% We learn \(F\), \(\pi\), and \(g\) to minimize task loss while penalizing information acquisition:
% \[
% \min_{F,\pi,g}\;
% \mathbb{E}_{(E,q,y)\sim\mathcal{D}}
% \left[
% \mathbb{E}_{\pi}\left(
% \ell(\hat{y},y) + \lambda\sum_{t=1}^{\tau-1} c(a_t)
% \right)
% \right],
% \]
% where \(c(a_t)\) captures cost (e.g., latency/compute) of acquiring observation \(o_{t+1}\), and \(\lambda\ge 0\) trades off accuracy and cost.

\subsection{Active Exploration Baseline}
We design the baseline agent to be one of the simplest possible instantiations of our agentic framework. The agent operates within an action space \(\mathcal{A} = \mathcal{A}_{\text{query}} \cup \{\textsc{Stop}\}\), where \(\mathcal{A}_{\text{query}} = \{\textsc{Query}\}\). Initially, \(\mathcal{F}_{\text{init}}\) just proposes a first action based on the question (without an expensive video-wide preprocessing step). At each iteration, it selects either \textsc{Stop} (terminating exploration to provide the final answer) or \textsc{Query(n)}, where n is a valid frame index. For a given query, the observation operator \(\mathcal{O}\) simply extracts the requested frame along with the surrounding transcript obtained using Whisper~\citep{radford2022robustspeechrecognitionlargescale}. The state update \(F\) invokes the backbone with the extracted RGB frame and transcript as well as the question and interaction history. % \xw{Are these extracted frames processed by the backbone or are they stored as RGB values? We might want to introduce the backbone in the formulation such as where it is used.}\tw{Clarified the manuscript.} 
In \(F\), the backbone outputs its reasoning along with the next action, both of which are appended to the interaction history. To prevent context overflow, this history is periodically compressed into a concise text summary using the backbone model. Notably, unlike many previous approaches, we do not temporally downsample the video before exploration; the agent retains access to the full original video and may query any frame at native temporal resolution. The full application of this approach to the above-defined formalism can be found in Appendix~\ref{sec:appendix_active_exploration_baseline}.

\subsection{SAVE (Simple Agentic Video Explorer)}
\label{sec:save_description}

While the simple baseline queries individual frames, SAVE expands the query action space to
\(
\mathcal{A}_{\text{query}} = \{\textsc{QueryRange}(s,e)\}
\), where \(s\) and \(e\) denote the start and end frame indices of a requested temporal interval. %\xw{how about using $t_s$ and $t_e$ for consistency?}\tw{I think it's fine here since we usually use $t$ for timestamps or agent iterations?} 
This allows the agent to request temporally contiguous regions of the video rather than isolated frames. We found that the lack of temporal context is a primary bottleneck of the minimal baseline, motivating this extension. Given the requested range \([s,e]\), the observation operator applies scene detection using \texttt{PySceneDetect}~\citep{castellano_pyscenedetect} and returns the middle frame of each detected scene segment. %\xw{If I understand it correctly, each scene segment is then represented by its middle frame? Is this what the PySceneDetect is doing or does it do something else as well?}\tw{Yes, that's what PySceneDetect is doing. To be precise, it outputs the start and end frame indices from which just take the mean to obtain the scene middle frame. Not sure if we need this implementation detail in text.}
The scene-detection granularity is selected automatically as a function of the queried range length. If the initial scene detection threshold yields too few scene segments, we progressively decrease the detection threshold until sufficient temporal coverage is obtained. %\xw{I am a bit lost here. What detection threshold is it here? It might be an misunderstanding because of the missing information about PySceneDetect.}\tw{PySceneDetect exposes a detection threshold which lets us control how much change there needs to be between frames for PySceneDetect to consider it a new scene. Worth adding this clarification here?}
If scene detection still fails to produce enough segments, we supplement the middle frames of detected scenes with uniformly sampled ones over the requested interval. SAVE additionally adapts the visual resolution of returned observations to the queried temporal scale: narrow ranges are processed at higher resolution to preserve fine-grained visual details, while broader ranges use default-resolution observations. This allows the agent to trade temporal coverage against visual detail without explicitly selecting these parameters itself. Additional settings are detailed in Appendix~\ref{sec:hyperparameters}. The full agent prompts and the summarized Algorithm~\ref{alg:save} are provided in Appendix~\ref{sec:prompts}. Figure~\ref{fig:save-scrabble} illustrates a SAVE trajectory on Minerva. %The algorithm is summarized in Algorithm~\ref{alg:save}.

\begin{figure}[t]
    \centering
    \includegraphics[width=0.8\linewidth]{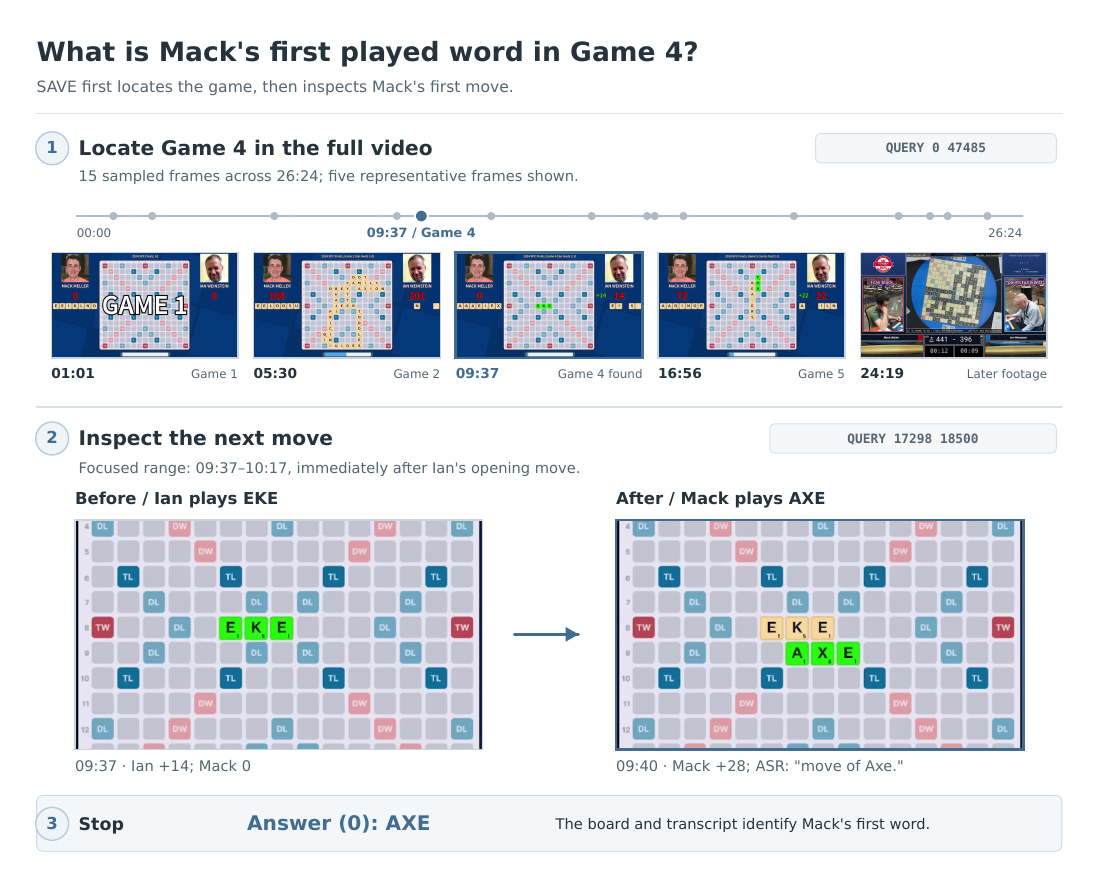}
    \caption{SAVE exploring Minerva video. A video-wide query
    locates Game 4 (five of 15 sampled frames shown). A focused query
    distinguishes Ian's opening EKE from Mack's response AXE, which
    the transcript also confirms. SAVE stops with the correct answer.}
    \label{fig:save-scrabble}
\end{figure}

\section{Experiments}

To the best of our knowledge, our study provides the most comprehensive controlled evaluation to date of contemporary agentic video reasoning methods under shared frontier backbones. We evaluate a diverse set of frameworks using both Gemini 3~\citep{gemini3_2026} and Gemma 4~\citep{gemma4}, while explicitly accounting for computational efficiency. By holding the principal multimodal backbone fixed within each comparison, this evaluation substantially reduces confounding from differences in underlying model capability and enables a more direct assessment of the accuracy--efficiency trade-offs of different exploration strategies.

\subsection{Experimental Setup}

\paragraph{Datasets.}
We evaluate on three complementary benchmarks: \textbf{Minerva}~\citep{nagrani2025minerva}, which evaluates multi-step reasoning over videos ranging from 2 minutes to over 1 hour; \textbf{EgoSchema}~\citep{mangalam2023egoschema}, which evaluates temporal, causal, and commonsense reasoning over three-minute egocentric videos; and \textbf{VSIBench}~\citep{yang2024think}, which evaluates visual-spatial reasoning. Due to the size of the VSIBench dataset we use the subsampling strategy from the original paper to obtain \emph{VSIBench (tiny)}, which contains 400 samples in total, 50 from each question type. 

\textbf{Baselines.} We run our evaluation on two specific backbones: Gemini 3 Flash~\citep{gemini3_2026} and Gemma 4~\citep{gemma4}. %\xw{We will probably get similar requests on using more models such as GPT and Qwen.}\tw{Yes, I think the two here give us high confidence in our assessment, much better than just one, but I can start evaluating on Qwen for the rebuttal.}
Our proposed method, SAVE, and the simple active exploration baseline are compared against the respective zero-shot VLM baseline and several popular agentic models: VideoAgent~\citep{VideoAgent}, VideoTree~\citep{wang2025videotree}, VideoCuriousAgent~\citep{yang2025vca}, VideoLucy~\citep{zuo2025videolucy}, DeepVideoDiscovery~\citep{zhang2025deep} and A.I.R.~\citep{zou2026air}. We select these specific baselines because they represent either current state-of-the-art models or conceptually significant exploration paradigms. Additionally, to verify that SAVE’s multi-step exploration paradigm provides benefits beyond one-shot query-conditioned localization, we include \emph{SAOS} (Scene-aware one-shot), which restricts SAVE to a single visual query.

\textbf{Evaluation Details.} For each method, we strictly adhere to the official author implementations where available, modifying only the multimodal backbone and the associated prompting logic (e.g. tool calls) to support the standardized backbone. %\xw{Are frontier models and backbones the same thing? Otherwise what is the standardized frontier model?}\tw{Yes, in this setup frontier models and backbones are the same thing, clarified the manuscript.}. 
In cases where official code is not publicly accessible, we reimplement the frameworks following the descriptions provided in their respective original publications. For hyperparameters, we follow the original authors’ recommendations wherever possible. All hyperparameters can be found in Appendix~\ref{sec:hyperparameters}. We detail and justify all baseline prompts we used in Appendix~\ref{appendix:baseline_prompts}. For the Gemini~3 native baseline, we utilize the API's default settings along with its built-in audio processing capabilities. We further attempt to include MR Video \citep{pang2025mrvideo} and AVP~\citep{wang2025activevideoperceptioniterative} in our evaluation. For MR Video we find its preprocessing requirements to be computationally prohibitive, exceeding 3M tokens for the first video in the Minerva dataset. AVP, meanwhile, is not fully backbone-agnostic: its released implementation relies on Gemini's native-video interface to execute temporal sampling and resolution-controlled observations, functionality that is not uniformly available across frontier backbones. Moreover, Gemini's native-video pipeline uses a different visual tokenization scheme from the image-frame inputs used by the other evaluated agents and directly processes the video's audio, whereas our standardized baselines receive Whisper-generated transcript snippets. These differences prevent a directly comparable backbone-standardized evaluation without materially modifying AVP's observation mechanism. We enable thinking for all baselines for evaluation on the Gemini backbone, but disable it when running the Gemma evaluation to test performance on both settings. We always use Gemini-2.5-Flash-Lite for  question-agnostic preprocessing steps requiring a VLM. To ensure consistent access to speech information, every non-native baseline receives the same Whisper-generated ASR transcript snippets surrounding the frames it requests in all evaluations. Further details on the evaluation protocol can be found in Appendix~\ref{sec:evaluation_protocol}.
To support future research and ensure full reproducibility, we will release all reimplementation code.

\begin{table*}[t]
\centering
\small
\setlength{\tabcolsep}{3.0pt}
\renewcommand{\arraystretch}{1.08}
\caption{
Performance and inference efficiency across video-reasoning benchmarks and backbones.
Each Gemini~3~Flash cell reports \emph{performance / total inference cost}, while each Gemma~4 cell reports \emph{performance / average tokens per sample}.
Performance is accuracy for Minerva and EgoSchema and the mean of the eight task-specific metrics for VSIBench. The bottom three rows show our own methods.
SAVE achieves the highest performance among evaluated prior agentic methods in five of six backbone--benchmark settings, while generally using substantially fewer resources than more elaborate agentic baselines.
Bold indicates the highest performance in each column.
See Appendix~\ref{sec:evaluation_protocol} for evaluation subsets, metrics, and cost accounting.
}
\label{tab:compressed_results}

\begin{tabular}{@{}lccc@{\hspace{6pt}}ccc@{}}
\toprule
&
\multicolumn{3}{c}{\textbf{Gemini 3 Flash} (\emph{Perf. / Cost})} &
\multicolumn{3}{c}{\textbf{Gemma 4} (\emph{Perf. / Tokens})} \\

\cmidrule(lr){2-4}
\cmidrule(lr){5-7}

\textbf{Model} &
\textbf{Minerva} &
\textbf{Ego.} &
\textbf{VSI} &
\textbf{Minerva} &
\textbf{Ego.} &
\textbf{VSI} \\
\midrule

Native
    & 64.8 / \$80.5
    & 75.4 / \$5.75
    & \textbf{59.2} / \$10.5
    & 51.6 / 15.9K
    & 72.4 / 16.0K
    & 49.1 / 15.9K \\

VideoTree
    & 50.4 / \$93.4
    & 73.1 / \$31.1
    & 45.5 / \$20.2
    & 51.6 / 36.4K
    & 71.1 / 34.4K
    & 42.8 / 24.6K \\

VideoAgent
    & 38.5 / \$34.7
    & 64.2 / \$4.76
    & 40.6 / \$10.6
    & 42.3 / 22.3K
    & 67.7 / 5.1K
    & 44.5 / 26.2K \\

VideoLucy
    & 56.5 / \$154.2
    & 72.7 / \$27.4
    & 48.2 / \$35.4
    & 58.4 / 72.0K
    & 73.2 / 103.6K
    & 43.8 / 63.4K \\

VideoCuriousAgent
    & 51.7 / \$235.8
    & 72.3 / \$51.8
    & 45.6 / \$61.2
    & 36.4 / 26.4K
    & 67.7 / 9.9K
    & 41.4 / 24.1K \\

DeepVideoDiscovery
    & 52.0 / \$30.5
    & 69.0 / \$11.5
    & 51.9 / \$5.21
    & 52.5 / 72.0K
    & 71.5 / 122.2K
    & 49.1 / 34.6K \\

A.I.R.
    & 55.0 / \$39.0
    & 73.9 / \$12.8
    & 49.9 / \$9.88
    & 51.9 / 22.1K
    & \textbf{74.3} / 23.0K
    & 45.7 / 20.2K \\

\midrule

SAOS
    & 57.5 / \$13.1
    & 73.1 / \$5.46
    & 50.0 / \$4.28
    & 52.5 / 8.9K
    & 70.5 / 9.9K
    & 43.5 / 8.6K \\

Active Exploration
    & 52.8 / \$38.1
    & 70.0 / \$3.89
    & 47.2 / \$9.40
    & 52.6 / 30.4K
    & 65.3 / 10.7K
    & 45.2 / 31.0K \\

SAVE
    & \textbf{66.4} / \$32.0
    & \textbf{76.0} / \$6.64
    & 57.1 / \$7.80
    & \textbf{62.4} / 19.7K
    & 72.4 / 9.4K
    & \textbf{52.6} / 11.1K \\

\bottomrule
\end{tabular}
\end{table*}

\subsection{Results}
All results are summarized in Table~\ref{tab:compressed_results}. The full breakdown of input and output token counts is provided in Appendix~\ref{sec:full_results}. Qualitative case studies can be found in Appendix~\ref{sec:case_studies}.

\textbf{Existing agentic methods do not consistently improve strong video backbones.}
To our surprise, across both Gemini~3 Flash and Gemma~4, agents do not consistently improve video reasoning performance, irrespective of the evaluated framework. With Gemini~3 Flash, none of the evaluated prior agentic methods outperforms native inference on Minerva, EgoSchema, or VSIBench, while several incur substantially greater inference cost. Remarkably, despite being restricted to a single visual query, SAOS achieves 57.5\% on Minerva, outperforming \emph{all evaluated agentic frameworks}, while being significantly cheaper overall than each agentic baseline. The same qualitative trend holds with Gemma~4: methods that improve over native inference often do so at substantially higher token cost. These results suggest that much of the apparent benefit of agentic pipelines diminishes once the underlying multimodal backbone is standardized.

\textbf{A minimal exploration policy is already competitive.}
We next investigate what makes an effective video agent: Our deliberately simple Active Exploration Baseline remains competitive with substantially more elaborate methods, reaching 52.8\% on Minerva. Its evaluation on Minerva costs less than all other agentic baselines apart from DVD and VideoAgent, both of which underperform this simple baseline. We found this result surprising given the recent trend in video reasoning toward increasingly elaborate agentic frameworks, including hierarchical memories, multi-stage retrieval pipelines, and specialized tool-use mechanisms~\citep{zhang2025deep,yin2026videoarmagenticreasoninghierarchical,zuo2025videolucy}. %\xw{This makes me wonder what other important benchmarks are used in the other studies and why we are not using the important ones.}\tw{It's quite fragmented, we do evaluate on EgoSchema which two of the cited approaches also evaluate on, but it's not directly comparable to our numbers since they use other proprietary models (GPT). One other important benchmark is VideoMME we could think about evaluating on it as well.}

\textbf{SAVE improves the accuracy--efficiency trade-off.}
SAVE consistently improves upon the minimal baseline, while still following a deliberately simple design. With Gemini~3 Flash, SAVE outperforms all evaluated prior agentic methods on Minerva, EgoSchema, and VSIBench. On Minerva, SAVE's performance is comparable to native inference at 40\% of its total inference cost. % \xw{I usually try to avoid repeat the results that people can read from the table directly, instead, we could write reletive improvements.}\tw{Done.}
Overall SAVE outperforms \emph{every prior agentic framework} in 5 out of 6 backbone--dataset settings. In the Gemini~3~Flash evaluation, SAVE achieves lower inference cost than all but one of the evaluated agentic baselines on each benchmark. The fact that SAVE matches, and in some settings outperforms, the corresponding backbone provides evidence that native inference is not always preferable to a well-designed agentic formulation, which motivates our analysis in Section~\ref{sec:comparison}.

\textbf{Statistical Significance and Comparison to Existing Results.} For benchmarks evaluated using per-example accuracy, we assess statistical significance using paired, two-sided tests with Holm correction for multiple comparisons: video-clustered permutation tests for Minerva, where multiple questions may correspond to the same video, and exact McNemar tests for EgoSchema.
In particular, SAVE significantly outperforms all agentic baselines on Minerva in both the Gemma and Gemini~3~Flash evaluations. The full results can be found in Appendix~\ref{sec:statistical_significance}. As further validation of our baseline reimplementations, we compare our results against previously reported results from related work in Appendix~\ref{sec:result_validation}.

\subsection{Architecture Ablations}

To understand the impact of our proposed enhancements, we systematically ablate the components of SAVE, building upon the simple Active Exploration Baseline. The results of this progressive ablation study across all datasets are presented in Table~\ref{tab:ablations-stats}.

Starting from the simple baseline, our initial analysis reveals that context breadth is the primary bottleneck. To address this, we transition from querying isolated frames to sampling entire scenes (\textbf{SAVE w/o HQ and AD}). In this setup the model chooses the granularity (low, medium, or high), which controls the scene detection thresholds. This substantially improves performance across all benchmarks, indicating that context breadth is a major component of successful video reasoning. Notably, performance on EgoSchema saturates after this improvement, suggesting that this is the major bottleneck on that particular benchmark~\footnote{An explanation for this is that EgoSchema tends to require less attention to small details than Minerva or VSIBench, because broad global context is usually sufficient for a correct answer.}. Certain tasks, however, require resolving precise visual details where perception becomes the limiting factor. To compensate, we introduce high-quality frames specifically for queries targeting fine temporal granularities (\textbf{SAVE w/o AD}). Consistent with our prior observation, performance increases on both Minerva and VSIBench. Finally, we observe that the model occasionally struggles in querying scenes at the right granularity. For example, this occurs in fine-grained motion sequences where standard scene detection fails to isolate meaningful semantic boundaries. To mitigate this, we introduce our adaptive threshold mechanism as described in Section~\ref{sec:save_description}. Its addition yields our final formulation (SAVE), which achieves the strongest overall performance across the three benchmarks, obtaining the highest scores on Minerva and VSIBench while remaining competitive on EgoSchema.

% \begin{table}[h]
% \centering
% \caption{Comparison of accuracy and token efficiency across video reasoning models on the LVBench dataset.}
% \label{tab:lvbench-stats}
% \begin{tabular}{lcccc}
% \toprule
% Model & Accuracy (\%) & Input Tokens & Output Tokens & Total Tokens \\
% \midrule
% Gemini 3 Flash Native & -- & -- & -- & -- \\
% VideoTree             & -- & -- & -- & -- \\
% VideoAgent            & -- & -- & -- & -- \\
% VideoLucy             & -- & -- & -- & -- \\
% VideoCuriousAgent     & -- & -- & -- & -- \\
% DeepVideoDiscovery    & -- & -- & -- & -- \\
% \midrule
% Active Search Baseline          & -- & -- & -- & -- \\
% \bottomrule
% \end{tabular}
% \end{table}

\begin{table}[t]
\centering
\small
    \caption{
Progressive ablation of SAVE. Replacing isolated-frame queries with scene-level range queries yields the largest initial gain across all three benchmarks. High-quality (HQ) observations for narrow ranges further improve Minerva and VSIBench, while adaptive scene detection (AD) gives the strongest performance on these two benchmarks; EgoSchema largely saturates after the scene-level extension. Bold and underline denote the best and second-best values within each dataset, respectively (higher is better for performance; lower is better for token counts).
}
\label{tab:ablations-stats}
\setlength{\tabcolsep}{3pt} % default is 6pt

\begin{tabular}{llcccc}
\toprule
Dataset & Model & Acc./Ave. (\%) & Input Tokens & Output Tokens & Total Tokens \\
\midrule
\multirow{4}{*}{Minerva} & Active Exploration Baseline & 52.8 & \textbf{29.4K} & 4.2K & \textbf{33.6K} \\
& SAVE w/o HQ and AD & 62.7 & \underline{36.2K} & \underline{2.5K} & \underline{38.7K} \\
& SAVE w/o AD & \underline{63.3} & 40.5K & 2.8K & 43.3K \\
& SAVE & \textbf{66.4} & 38.3K & \textbf{1.5K} & 39.7K \\
\midrule
\multirow{4}{*}{VSIBench (tiny)} & Active Exp. Baseline & 47.2 & \textbf{25.8K} & 3.5K & \textbf{29.3K} \\
& SAVE w/o HQ and AD & 51.3 & 33.2K & 3.0K & 36.2K \\
& SAVE w/o AD & \underline{52.5} & 34.2K & \underline{2.7K} & 36.9K \\
& SAVE & \textbf{57.1} & \underline{28.5K} & \textbf{1.7K} & \underline{30.3K} \\
\midrule
\multirow{4}{*}{EgoSchema} & Active Exploration Baseline & 70.0 & \textbf{9.1K} & 1.1K & \textbf{10.2K} \\
& SAVE w/o HQ and AD & \textbf{77.0} & \underline{16.9K} & \underline{0.9K} & \underline{17.8K} \\
& SAVE w/o AD & 75.6 & 17.8K & 1.0K & 18.8K \\
& SAVE & \underline{76.0} & 21.6K & \textbf{0.9K} & 22.5K \\
\bottomrule
\end{tabular}
\end{table}

\section{When does Agentic Video Reasoning Help?}
\label{sec:comparison}

\begin{figure}
    \centering
    \includegraphics[width=\columnwidth]{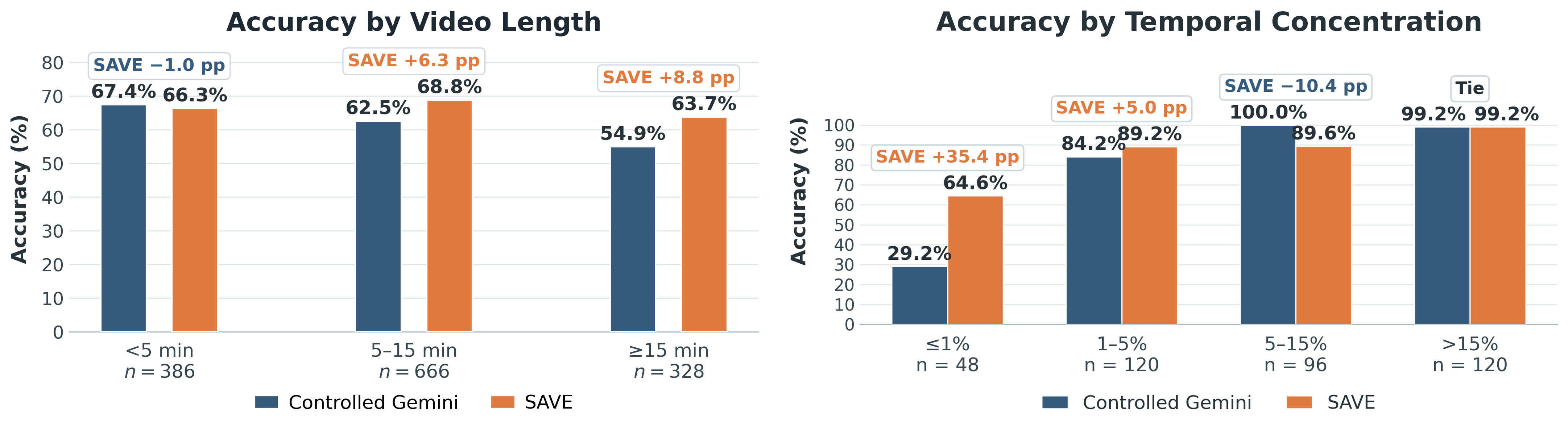}
    \vspace{-5mm}
    \caption{
Controlled comparison with Gemini~3~Flash under approximately matched input-token budgets. Left: on Minerva~\citep{nagrani2025minerva}, SAVE is comparable to native inference for videos under 5 minutes, but its advantage grows with video length. Right: on our synthetic benchmark, SAVE provides the largest gain when answer-relevant events occupy only a small fraction of the video; the advantage vanishes as the required temporal pattern becomes broadly represented. Bars show accuracy; $n$ denotes the number of samples in each bin.
}
    \label{fig:controlled_experiment}
\end{figure}

Motivated by the fact that SAVE is competitive with, and in several settings outperforms, native processing, we systematically
investigate when the agentic formulation is preferable to native inference and vice-versa. Our results indicate that
two key factors are \emph{video length} and \emph{temporal concentration of answer-relevant evidence}. 

\paragraph{Video Length.} Dense processing remains practical for short videos, leaving limited scope for active exploration to improve accuracy while introducing substantial overhead. Selective inspection becomes increasingly valuable as videos grow longer, because dense processing must distribute computation across the full video and can incur substantial memory and compute costs. In light of this, we hypothesize that under equal token budgets active exploration becomes increasingly useful as the duration of the video grows. We test this hypothesis % \xw{What is the hypothesis?}\tw{Clarified the manuscript.} 
on Minerva, whose videos range from under two minutes to over an hour. Using Gemini~3~Flash for both approaches, we calibrate the sampling rate of direct full-video inference to approximate SAVE's per-sample input-token consumption while holding the generation configuration fixed. We verify that this calibration results in closely matched total inference costs: \$31.45 for Gemini and \$31.70 for SAVE. SAVE significantly outperforms Controlled Gemini (66.88\% vs.\ 62.03\%; video-clustered permutation test, p=0.0055). As shown in Figure~\ref{fig:controlled_experiment}, the methods perform comparably on videos shorter than five minutes (66.32\% vs.\ 67.36\%; p=0.776). SAVE's advantage then grows to 6.31 percentage points on 5--15 minute videos (68.77\% vs.\ 62.46\%; p=0.009) and 8.84 points on videos lasting at least 15 minutes (63.72\% vs.\ 54.88\%; p=0.001).

\paragraph{Temporal Concentration of Evidence.} Duration alone, however, does not fully determine whether active exploration is beneficial. Even in short videos, answering a question may require resolving a temporal pattern that unfolds over only a small fraction of the video's duration. Under a fixed input-token budget, uniform sampling spreads observations across the entire video and may therefore sample too sparsely to capture such fine-grained temporal structure. An agentic method can instead adapt its sampling granularity, ``zooming in'' to inspect candidate intervals at higher temporal resolution.

To isolate this effect, we introduce a synthetic video question-answering benchmark with 384 samples and video durations ranging from 30 seconds to 10 minutes. Each video contains a precisely defined answer-relevant event that repeats continuously throughout the video, making localization trivial. This construction allows us to study whether adaptive sampling helps when the temporal resolution required to resolve an event becomes fine relative to the overall video duration, separately from the difficulty of localizing the event itself. Details are provided in Appendix~\ref{sec:synthetic_benchmark}. We define \emph{temporal concentration} as the fraction of the video occupied by one sufficient event, i.e., the event duration divided by the video duration. Smaller values indicate that a sufficient event unfolds over a smaller fraction of the video and therefore requires finer temporal resolution relative to the overall video duration.

Across all 384 common samples, SAVE and Controlled Gemini (approximately matching SAVE's token consumption) achieve 89.32\% and 85.94\% accuracy, respectively. However, as shown in Figure~\ref{fig:controlled_experiment}, when an individual event lasts at most 1\% of the video, SAVE significantly outperforms Controlled Gemini by 35.42 percentage points (64.58\% vs.\ 29.17\%, \(p = 0.0015\)). As the event duration increases relative to the video duration, uniform sampling becomes increasingly likely to capture the relevant temporal pattern at sufficient resolution, reducing the benefit of adaptive refinement and making the overhead of the agentic formulation less worthwhile. To verify that SAVE obtains its advantage through iterative adaptive sampling, we also evaluate SAOS on the synthetic dataset. SAOS performs substantially worse overall (46.1\% vs. 89.3\%). SAVE's lead over SAOS generally increases as the sequence of events becomes more temporally concentrated: 24.2 percentage points for (>15\%), 36.5 for (5\text{--}15\%), and 69.2 for (1\text{--}5\%), before narrowing to 39.6 points for ($\leq$ 1\%). We further verify that SAVE's advantage on this benchmark is not driven by scene detection: PySceneDetect identifies no scene transitions in any of the synthetic samples, regardless of the threshold used by SAVE. Additionally, we confirm that Controlled Gemini closely matches SAVE's inference cost, with total costs of \$4.73 and \$4.63, respectively.

\paragraph{Implications.} Together, these experiments suggest a simple principle: active exploration is most useful when a fixed processing budget would otherwise represent answer-relevant evidence too sparsely. Long-form video is therefore a particularly promising regime, since temporal coverage becomes increasingly expensive while fine-grained events may still need to be resolved. This interpretation is consistent with the direction of recent work: The temporal horizon targeted by agentic methods has shifted toward progressively longer videos. Early approaches such as VideoAgent~\citep{VideoAgent} evaluated on minute-scale benchmarks such as NExT-QA~\citep{xiao2021nextqa}, while VideoTree~\citep{wang2025videotree} already extended evaluation to the long split of Video-MME~\citep{fu2025videomme}, with videos averaging roughly 44 minutes. This shift is also reflected in the benchmark landscape: LVBench~\citep{lvbenchwang2025} explicitly targets extreme long-video understanding, increasing the average video duration to approximately 68 minutes, compared with roughly 10 minutes for earlier long-video benchmarks. More recent systems such as DeepVideoDiscovery~\citep{zhang2025deep} explicitly target information-dense hour-long videos, while the work of~\citet{rege2026agenticverylong} pushes the setting further toward longitudinal video streams spanning multiple days.

\section{Limitations}

Our study is limited to a finite set of contemporary agentic frameworks, two strong multimodal backbones, and three video-reasoning benchmarks. Standardizing heterogeneous methods requires adapting backbone calls, prompts, and tool interfaces, which may affect their behavior despite our efforts to preserve each method's algorithmic structure. We also do not evaluate on an ultra-long benchmark such as LVBench~\citep{lvbenchwang2025}, where expensive video-wide preprocessing may become more advantageous.

Finally, our efficiency analysis primarily measures token consumption and inference cost. These capture the dominant model-side computational expense, but do not fully reflect all factors relevant in practice. We do not include end-to-end latency, as in our evaluation setting it is highly noisy and strongly influenced by external factors unrelated to the method itself, such as public API rate limiting, transient outages, network variability and heterogeneous hardware setups. We therefore view token usage and inference cost as the more reliable basis for controlled comparison.

\section{Conclusion}

We presented a unified view of agentic video reasoning and conducted a controlled evaluation of contemporary methods under shared frontier backbones. Our results show that increasingly elaborate agentic frameworks do not reliably outperform native video inference, while a minimal backbone-driven exploration strategy remains highly competitive. Building on this observation, SAVE outperforms all evaluated prior agentic frameworks in five of six backbone--benchmark settings. Moreover, our analysis suggests that agentic video reasoning is most useful when answer-relevant evidence requires fine temporal resolution relative to the overall video duration, or when exhaustive native processing is costly. These findings motivate future agentic video reasoning research to focus primarily on long-form video understanding.

\subsection*{AI use statement}
AI tools were used to support literature research, assist with the implementation of experimental baselines, and check these implementations for faithfulness to the corresponding papers and released code. They were also used for proofreading and improving the clarity of the manuscript, as well as for supporting data analysis. All methodological decisions, experimental design choices, implementation decisions, and interpretations of the results were made and verified by the authors.

% TODO

\clearpage

\bibliographystyle{iclr2027_conference}
\bibliography{references}

\clearpage
\appendix

\section{Implementation details}
\label{sec:implementation_details}

\subsection{Evaluation Protocol}
\label{sec:evaluation_protocol}

\paragraph{Gemma Evaluation} For the Gemma~4 evaluation, which we run locally, we are restricted by the available GPU memory and therefore cannot process an arbitrary number of frames in a single model call. We consequently limit the number of frames provided to Gemma~4 to at most 60 per call for all evaluated methods. For the native Gemma~4 baseline, we uniformly sample 60 frames from the full video. For agentic methods, this same per-call limit is applied to every intermediate reasoning or observation step, ensuring that all methods operate under the same model-level memory constraint. We emphasize that the memory requirements are an explicit disadvantage of native inference.

\paragraph{Evaluation Subsets} For all experiments, a small portion of videos in each dataset is either blocked by the Gemini API's safety filters or no longer publicly available. For Minerva specifically a small portion of videos exceed Gemini's context length using the default settings. Consequently, we evaluate on the remaining accessible subsets. Specifically, on Minerva we evaluate 1359 out of 1418 samples, and on EgoSchema we evaluate 491 out of 500 samples from the labeled subset. For VSIBench, due to the high computational overhead of many-agent comparisons, we evaluate 400 samples (taking the first 50 from each question type), following the subsampling strategy used in the original paper~\citet{yang2024think}. For the controlled Minerva experiment in Section~\ref{sec:comparison}, we evaluate on the 1,380 samples for which both SAVE and the token-matched native baseline produce valid outputs. This subset differs from the 1,359-sample subset used in the main evaluation because the controlled native baseline processes fewer uniformly sampled frames, causing fewer videos to exceed Gemini’s context limit. All comparisons within each experiment are performed on the corresponding common valid subset.

\paragraph{Metrics}
For our evaluation on EgoSchema and Minerva, we report the standard Accuracy metric. For VSIBench, following the original evaluation protocol, we report the average score across its eight question types. Numerical-answer tasks (object counting, object size, room size, and absolute distance) are evaluated using Mean Relative Accuracy (MRA), which measures numerical accuracy across a range of relative-error tolerances. Multiple-choice tasks (relative distance, relative direction, route planning, and appearance order) are evaluated using standard accuracy. The overall VSIBench score is the unweighted average across these eight task-level scores.

\paragraph{Cost Accounting}
As Gemini~3~Flash is a proprietary model accessed through an API, we use monetary inference cost, based on the provider's API pricing, as the primary efficiency metric. When additionally reporting token counts for Gemini~3~Flash, thinking tokens are included as output tokens. For methods that require question-agnostic preprocessing, the corresponding preprocessing cost is incurred only once per video and amortized across all questions associated with that video. Gemma~4 is an open-weight model that we evaluate locally, so there is no corresponding API-based inference cost; we therefore report total model tokens as the primary efficiency metric. Since Gemma~4 is evaluated without thinking enabled, its output-token counts correspond only to the visible generated output. For readability, we always report per-sample averages for token counts and totals across the full evaluation for monetary inference cost

\subsection{Hyperparameters}
\label{sec:hyperparameters}
All Gemini-based baselines used \texttt{gemini-3-flash-preview} with Decord frame decoding.

\paragraph{VideoAgent.}
VideoAgent initially sampled 5 uniformly spaced frames and allowed up to 4 refinement iterations. Additional frames during refinement were selected using CLIP-based frame retrieval. The final answer was taken from the last model response.

\paragraph{VideoTree.}
We sampled frames at 1 FPS and used EVA-CLIP embeddings with cosine $k$-means. We set the initial number of clusters to 8, maximum breadth to 32, maximum depth to 3, branch width to 4, and the high-relevance stopping threshold to 4. Gemini calls used temperature 0.0, a thinking-budget cap of 256 tokens, and caption chunks of 10 frames.

\paragraph{VideoCuriousAgent.}
Each exploration step sampled 4 frames from the selected segment. We used a memory buffer of 16 frames and capped exploration at 5 iterations. Gemini calls used temperature 0.5.

\paragraph{VideoLucy.}
Coarse memory captioning used \texttt{gemini-2.5-flash-lite}; fine/ultra-fine captioning and all reasoning calls used \texttt{gemini-3-flash-preview} or \texttt{gemma-4-31b-it}. We used a thinking-budget cap of 2048 tokens and at most 5 backtracking iterations. Coarse memory was sampled at 1 FPS and fine/ultra-fine clips at 2 FPS. For Minerva, the coarse temporal scope was set by video duration to 30/60/100/200 seconds for videos shorter than 600/1200/3600 seconds or longer, respectively; the fine scope was 5/5/10/10 seconds. The ultra-fine scope was 1 second.

\paragraph{DeepVideoDiscovery.}
The video database was built from 10-second clips sampled at 2 FPS. Preprocessing used \texttt{gemini-2.5-flash-lite} for clip captioning/registry merging and \texttt{gemini-embedding-001} with 3072-dimensional embeddings. The agent used \texttt{gemini-3-flash-preview} or \texttt{gemma-4-31b-it} with function calling. We capped interaction at 3 iterations, used a global-browse top-$k$ of 300, and allowed frame-inspection calls to include up to 50 frames.

\paragraph{SAVE.}
SAVE allowed the active vision agent to issue up to 30 visual queries before producing a final answer. Each query specified a frame-index range, and the scene-detection parameters were selected automatically based on the duration of the requested interval. For ranges of at most 10 seconds, we used a target of 6 scenes and \texttt{ContentDetector} thresholds of 12, 10, 8, and 6. For ranges of at most 60 seconds, we used a target of 12 scenes and thresholds of 20, 18, 16, 14, 12, and 10. For ranges of at most 8 minutes, we used a target of 25 scenes and thresholds of 30, 25, 20, 18, 16, 14, and 12. For longer ranges, we used a target of 6 scenes and thresholds of 40, 35, 30, 25, 20, 16, 12, and 10. For each query, PySceneDetect progressively reduced the detection threshold until the target number of scenes was reached or all configured thresholds had been evaluated. The environment returned the middle frame of each detected scene and uniformly subsampled the detected scenes to at most $K=40$ frames per query. If no threshold produced the target number of scenes, SAVE supplemented the scene-middle frames with uniformly sampled frames over the queried interval to reach the targeted number of scenes. 
Query ranges of at most $\delta=10$ seconds used high-resolution visual inputs, whereas longer ranges used default-resolution inputs. For each returned frame, SAVE additionally provided the local transcript from approximately one second before to one second after the corresponding timestamp. Across iterations, SAVE retained a text-only interaction history containing the queried ranges, relevant frame indices, transcript snippets, and the model's textual observation summaries. The history was compressed every $C=5$ exploration steps using the backbone with a thinking budget of 2048 tokens, while preserving answer-relevant evidence and temporal search anchors.

\subsection{Baseline Prompts}
\label{appendix:baseline_prompts}
We provide the prompts we use inside each baseline implementation along with any deviations from the original formulations. To ensure consistent access to speech information across methods, we append ASR-generated transcripts to the prompts below if applicable.

\subsubsection{VideoTree} 
\paragraph{Captioning prompt}

\begin{quote}
You are a precise video captioner. You will receive multiple frames from a single video together with their frame indices.

\textbf{Question context:}

\emph{[question]}

You \textbf{must} caption every frame with an index in
\[
[\text{\emph{frame indices}}]
\]
exactly once. Do not skip or duplicate any frame index.

For each frame, output one concise sentence describing the visible actions and objects relevant to the question.

Return only a JSON array containing exactly one object per frame, using the following schema:
\begin{verbatim}
[{"frame_idx": 123, "caption": "..."}, ...]
\end{verbatim}

Always use exactly the keys \texttt{frame\_idx} and \texttt{caption}. Do not include Markdown, additional keys, or any text outside the JSON array.
\end{quote}

The VideoTree repository~\footnote{https://github.com/Ziyang412/VideoTree} does not provide a captioning prompt as all the captions come pre-generated from LaViLA~\cite{zhao2023lavila}. We therefore introduce the prompt above to obtain one question-conditioned caption for each requested frame.

\paragraph{Relevance Scoring Prompt}

\begin{quote} You are presented with a textual description of a video clip consisting of \(N\) frame captions sparsely sampled from the video. Your goal is to answer a question about the video. Imagine the visual scene as vividly as possible to improve the accuracy of your response. After selecting an answer, assign a confidence score from \(1\) to \(100\), where \(1\) indicates low confidence and \(100\) indicates high confidence. Provide a concise, one-sentence explanation for your answer. If uncertain, select the answer that appears most likely to be correct. Additionally, assign a relevance score to each frame caption according to its relevance to answering the question. Each score must be one of \(\{1,2,3\}\), where \(1\) indicates low relevance and \(3\) indicates high relevance. Return exactly one relevance score per frame caption, preserving the order in which the captions are provided. If the question contains answer options, return the corresponding option identifier as the prediction, such as \texttt{A}, \texttt{B}, \texttt{C}, or a numeric identifier. If no explicit answer options are provided, return a short free-form answer. Never return \texttt{null}, \texttt{unknown}, \texttt{N/A}, or \texttt{-1}; when the exact answer cannot be determined, provide the most plausible estimate. \textbf{Description:} \emph{[frame captions]} \medskip \noindent\texttt{\#\#\#} \textbf{Transcript snippets:} \emph{[transcript]} \medskip \noindent\texttt{\#\#\#} \textbf{Question:} \emph{[question]} \medskip \noindent\texttt{\#\#\#} Return only a JSON object in the following format: \begin{verbatim} { 
"prediction": "...", 
"explanation": "...", 
"confidence": 1, 
"frame_relevance": [1, 2, 3] 
} \end{verbatim} \end{quote}

The original VideoTree prompt is tailored to EgoSchema: it assumes an egocentric video, uses dataset-specific caption conventions such as \#C and \#O, and restricts predictions to five lettered answer choices. We adapt the prompt for evaluation across a broader set of benchmarks by removing these dataset-specific assumptions and supporting each benchmark’s answer format. For a fair comparison, we also include the same ASR transcript provided to every evaluated method. Finally, because the evaluated backbones did not reliably follow the original free-form response format, we require structured JSON output to ensure robust and consistent parsing. These adaptations preserve the core relevance-scoring task of assigning a score from 1 to 3 to each selected caption.

\paragraph{Final Reasoning Prompt}

\begin{quote}
You are given a query-adaptive, coarse-to-fine set of frame captions from a long video. Use them to answer the question as accurately as possible.

\textbf{Transcript snippets:}

\emph{[transcript]}

\textbf{Frame captions in temporal order:}

\emph{[frame captions]}

\textbf{Question:}

\emph{[question]}

If the question provides answer options, return the corresponding option identifier as the prediction, such as \texttt{A}, \texttt{B}, \texttt{C}, or a numeric identifier. If the question requires a numeric answer and does not provide answer options, return only the number as the prediction, for example \texttt{3.5}.

Never return \texttt{null}, \texttt{unknown}, \texttt{N/A}, or \texttt{-1}. If the exact answer cannot be determined, provide the most plausible estimate.

Return only a JSON object with the following keys:
\begin{verbatim}
{
  "prediction": "option identifier or numeric answer",
  "explanation": "brief reasoning",
  "confidence": 1
}
\end{verbatim}

The value of \texttt{confidence} must be an integer from \(1\) to \(100\).
\end{quote}

Again we adapt VideoTree’s final-reasoning prompt to support multiple benchmarks rather than the EgoSchema-specific five-choice setting. The adapted prompt supports benchmark-specific option identifiers and numerical answers and presents the selected captions in temporal order. For a controlled comparison, each caption is annotated with its source timestamp, matching the temporal information available to the other evaluated methods and supporting questions involving event order, duration, or explicit temporal references. We also include the same ASR transcript supplied to every method. Finally, we require structured JSON output to ensure reliable parsing across model backbones.

\subsubsection{VideoLucy}
\paragraph{Coarse Captioning Prompt}
\begin{quote} Observe and understand the given video carefully. Describe its contents as comprehensively as possible in a smooth and coherent passage. Do not omit any details or prominent information. If the video contains any text, subtitles, text overlays, or voice-overs, explicitly describe their content in detail. \end{quote}

This prompt is identical to the one from the VideoLucy repo\footnote{https://github.com/worldbench/VideoLucy}.

\paragraph{Coarse Answering Prompt}

\begin{quote}
The following provides a rough description of what is shown in the video during different time periods:

\textbf{Video descriptions:}

\emph{[memory]}

Note that since these descriptions are not very complete and detailed, some key information in the video segments of each time period may not all appear in these content descriptions.

This is an academic video understanding benchmark task. The following is the benchmark question to answer:

\textbf{Question:}

\emph{[question]}

Please read the given video content descriptions and the question in depth, and determine whether you can accurately answer the given question solely based on the currently provided descriptions.

If you can answer it with absolute confidence, please answer this question and provide the time periods you are referring to. The answer you provide must have completely and absolutely objective support in the video descriptions. Do not make inferences arbitrarily.

If you think the current content descriptions of the video are still insufficient to accurately answer the question, please do not answer it and give me your reason.

Please output in a strictly standardized dictionary format containing four key--value pairs:

\begin{itemize}
    \item \texttt{"Confidence"}: A boolean value. Set it to \texttt{True} if you are certain about the answer, and \texttt{False} if not.

    \item \texttt{"Answer"}: A string. This string must be enclosed in double quotes. When \texttt{"Confidence"} is \texttt{True}, fill in the correct answer; when \texttt{"Confidence"} is \texttt{False}, fill in \texttt{"No Answer"}. This must be exactly one number, either the number of the option, if options are given, or the numeric answer to the question, if no options are given.

    \item \texttt{"Time Period"}: A list. When \texttt{"Confidence"} is \texttt{True}, fill in the list with time periods corresponding to the answer, each in the format of a tuple \texttt{(start time, end time)}; when \texttt{"Confidence"} is \texttt{False}, fill in \texttt{"No Time"}.

    \item \texttt{"Reason"}: A string. This string must be enclosed in double quotes. Show me your reasoning about your judgment. You need to ensure and check that your reasoning must be able to absolutely support your answer.
\end{itemize}

Note that no additional comments should be added within the dictionary.
You must note that if an ordinal number appears in the provided question, in the vast majority of cases, you should not simply assume that this ordinal number represents the ordinal of the provided time period. You need to focus on understanding the specific meaning of this
ordinal number within the question based on all the content descriptions.
You \textbf{must} output a valid JSON dictionary.

Return \textbf{only} a valid JSON object.

The response must start with \texttt{\{} and end with \texttt{\}}.

Do not wrap the JSON in Markdown fences.

Do not add explanations before or after the JSON.

Use double quotes for all keys and string values.

Use JSON booleans \texttt{true}/\texttt{false}, not Python booleans \texttt{True}/\texttt{False}.

For \texttt{"Time Period"}, use JSON arrays such as \texttt{[[120, 150]]}, not tuples such as \texttt{[(120, 150)]} or strings such as \texttt{"120s-150s"}.
\end{quote}

For this prompt, in contrast to the original prompt, we again adjust the answer format to be dataset-agnostic, require a strict JSON output for reliability and introduce the academic benchmark framing to bypass Gemini API filters for certain Minerva questions.  

\paragraph{Initial Localization Prompt}
\begin{quote}
The following provides a rough description of what is shown in the video during different time periods:

\textbf{Video descriptions:}

\emph{[memory]}

Now, a question has been raised regarding this video:

\textbf{Question:}

\emph{[question]}

Please read the given video content descriptions and the question in depth.

Since most of these descriptions are rather rough and some detailed information is lost, my task is to try my best to find the time periods related to the given question and then provide more detailed descriptions of the video content during these time periods.

In order to assist me in completing this task, your task is to determine, based on the provided rough video descriptions, whether further observing the video content during two time periods would allow me to answer the given question with greater confidence.

If so, identify the time periods related to the question as accurately as possible and provide these relevant time periods so that I can review the corresponding video segments again, obtain more information, and answer the question more accurately.

For example, since they do not require an overall understanding of large video segments, the following questions can be answered more accurately by carefully re-observing two video segments:

\begin{enumerate}[label=(\roman*)]
    \item What color is Putin's tie between the interview with Antony Blinken and the interview with Marie Yovanovitch?
    \item How does the goalkeeper prevent Liverpool's shot from scoring at 81:38 in the video?
    \item Who smashes the magic mirror?
\end{enumerate}

In contrast, because they require an overall understanding of large video segments, it is difficult to answer the following questions more accurately by observing only two video segments:

\begin{enumerate}[label=(\roman*)]
    \item What happens in the second half of the game?
    \item What is the video about?
    \item Which places has the protagonist of this video visited in total?
\end{enumerate}

You should output a strictly standardized dictionary containing exactly three key--value pairs:

\begin{itemize}
    \item \texttt{"Flag"}: A Boolean value. If you are very confident that you can provide time periods according to the requirements above, set it to \texttt{True}. Otherwise, set it to \texttt{False}.

    \item \texttt{"Time Period"}: A list. If \texttt{"Flag"} is \texttt{True}, fill the list with the two most relevant time periods, each in the tuple format \texttt{(start time, end time)}. If \texttt{"Flag"} is \texttt{False}, return \texttt{"No Time Periods"}.

    \item \texttt{"Reason"}: A string enclosed in double quotation marks. Explain the reasons for the time periods you provided.
\end{itemize}

No additional comments should be added within the dictionary.

Return \textbf{only} a valid JSON object.

The response must start with \texttt{\{} and end with \texttt{\}}.

Do not wrap the JSON in Markdown fences.

Do not add explanations before or after the JSON.

Use double quotation marks for all keys and string values.

Use JSON Boolean values \texttt{true}/\texttt{false}, not Python Boolean values \texttt{True}/\texttt{False}.

For \texttt{"Time Period"}, use JSON arrays such as \texttt{[[120, 150]]}, not tuples such as \texttt{[(120, 150)]} or strings such as \texttt{"120s-150s"}.
\end{quote}

We preserve VideoLucy’s original localization-and-instruction prompt while stating its existing exclusion of previously searched periods more explicitly, as the API-based model occasionally revisited excluded intervals in preliminary Minerva runs. For cross-dataset evaluation, we additionally embed answer choices in the question and enforce valid JSON outputs for reliable inference through the Gemini API.

\paragraph{Localization and Instruction Agent Prompt}

\begin{quote}
The following provides a rough description of what is shown in the video during different time periods:

\textbf{Video descriptions:}

\emph{[memory]}

Note that since these descriptions are not very complete or detailed, some key information from the video segments corresponding to each time period may not appear in the content descriptions.

This is an academic video-understanding benchmark task. The following is the benchmark question to answer:

\textbf{Question:}

\emph{[question]}

Please read the provided video content descriptions and the question in depth, and determine whether you can accurately answer the question solely based on the currently provided descriptions.

If you can answer it with absolute confidence, answer the question and provide the time periods to which you are referring. Your answer must have complete and objective support in the video descriptions. Do not make arbitrary inferences.

If you think the current video descriptions are still insufficient to answer the question accurately, do not answer it and instead provide your reason.

Please output a strictly standardized dictionary containing exactly four key--value pairs:

\begin{itemize}
    \item \texttt{"Confidence"}: A Boolean value. Set it to \texttt{True} if you are certain about the answer and \texttt{False} otherwise.

    \item \texttt{"Answer"}: A string enclosed in double quotation marks. When \texttt{"Confidence"} is \texttt{True}, provide the correct answer. When \texttt{"Confidence"} is \texttt{False}, return \texttt{"No Answer"}. The answer must contain exactly one number: either the number of the selected option, if options are provided, or the numeric answer to the question, if no options are provided.

    \item \texttt{"Time Period"}: A list. When \texttt{"Confidence"} is \texttt{True}, fill the list with the time periods corresponding to the answer, with each time period represented as a tuple in the format \texttt{(start time, end time)}. When \texttt{"Confidence"} is \texttt{False}, return \texttt{"No Time"}.

    \item \texttt{"Reason"}: A string enclosed in double quotation marks. Explain the reasoning behind your judgment. Ensure that the reasoning provides complete and objective support for your answer.
\end{itemize}

Do not add any additional comments within the dictionary.

If an ordinal number appears in the provided question, you must note that, in the vast majority of cases, you should not simply assume that this ordinal number represents the ordinal position of a provided time period. Instead, focus on understanding the specific meaning of the ordinal number within the question based on all of the content descriptions.

You \textbf{must} output a valid JSON dictionary.

Return \textbf{only} a valid JSON object.

The response must start with \texttt{\{} and end with \texttt{\}}.

Do not wrap the JSON in Markdown fences.

Do not add explanations before or after the JSON.

Use double quotation marks for all keys and string values.

Use JSON Boolean values \texttt{true}/\texttt{false}, not Python Boolean values \texttt{True}/\texttt{False}.

For \texttt{"Time Period"}, use JSON arrays such as \texttt{[[120, 150]]}, not tuples such as \texttt{[(120, 150)]} or strings such as \texttt{"120s-150s"}.
\end{quote}

We preserve VideoLucy’s original localization-and-instruction prompt while making its existing exclusion of previously searched periods more explicit after having observed deviations from intended behavior on Minerva. Furthermore, we embed answer choices in the question (for cross-dataset evaluation), and enforce valid JSON outputs for reliable Gemini API inference.

\paragraph{Captioning Prompt}

\begin{quote} You should explicitly and in detail describe any visible subtitles, text overlays, or voice-overs in the video. In addition, follow the instruction below: \emph{[instruction]} \end{quote}

The intermediate captioning prompt is exactly identical to the one in the original repository.

\paragraph{Answering Prompt}

\begin{quote}
There is currently a video with a total duration of \emph{[video length]} seconds.

The following provides a general description of what is shown in the video during certain time periods:

\textbf{Video descriptions:}

\emph{[memory]}

This is an academic video-understanding benchmark task. The following is the benchmark question to answer:

\textbf{Question:}

\emph{[question]}

Please read the provided video content descriptions and the question in depth, and determine whether you can accurately answer the question solely based on the currently provided descriptions.

If you can answer it with absolute confidence, answer the question and provide the time periods of the video content to which you are referring. Your answer must have complete and objective support in the video descriptions. Do not make arbitrary inferences.

If you think the current video descriptions are still insufficient to answer the question accurately, do not answer it and instead provide your reason.

Please output a strictly standardized dictionary containing exactly four key--value pairs:

\begin{itemize}
    \item \texttt{"Confidence"}: A Boolean value. Set it to \texttt{True} if you are certain about the answer and \texttt{False} otherwise.

    \item \texttt{"Answer"}: A string enclosed in double quotation marks. When \texttt{"Confidence"} is \texttt{True}, provide the correct answer option. When \texttt{"Confidence"} is \texttt{False}, return \texttt{"No Answer"}. The answer must contain exactly one number: either the number of the selected option, if options are provided, or the numeric answer to the question, if no options are provided.

    \item \texttt{"Time Period"}: A list. When \texttt{"Confidence"} is \texttt{True}, fill the list with the time periods corresponding to the answer, with each time period represented as a tuple in the format \texttt{(start time, end time)}. When \texttt{"Confidence"} is \texttt{False}, return \texttt{"No Time"}.

    \item \texttt{"Reason"}: A string enclosed in double quotation marks. Explain the reasoning behind your judgment. Ensure that the reasoning provides complete and objective support for your answer.
\end{itemize}

Do not add any additional comments within the dictionary.

If an ordinal number appears in the provided question, note that, in the vast majority of cases, you should not simply assume that this ordinal number represents the ordinal position of a provided time period. Instead, focus on understanding the specific meaning of the ordinal number within the question based on all of the content descriptions.

Return \textbf{only} a valid JSON object.

The response must start with \texttt{\{} and end with \texttt{\}}.

Do not wrap the JSON in Markdown fences.

Do not add explanations before or after the JSON.

Use double quotation marks for all keys and string values.

Use JSON Boolean values \texttt{true}/\texttt{false}, not Python Boolean values \texttt{True}/\texttt{False}.

For \texttt{"Time Period"}, use JSON arrays such as \texttt{[[120, 150]]}, not tuples such as \texttt{[(120, 150)]} or strings such as \texttt{"120s-150s"}.
\end{quote}

We preserve VideoLucy’s original answering policy while framing each query as an academic benchmark task to mitigate false-positive Gemini API filtering observed on Minerva. We further standardize predictions as dataset-agnostic numeric outputs and require valid JSON to ensure reliable parsing across datasets.

\paragraph{Forced Answer Prompt}

\begin{quote}
There is currently a video with a total duration of \emph{[video length]} seconds.

The following provides a general description of what is shown in the video during certain time periods:

\textbf{Video descriptions:}

\emph{[memory]}

This is an academic video-understanding benchmark task. The following is the benchmark question to answer:

\textbf{Question:}

\emph{[question]}

Please read and understand the provided video content and question in depth. Strictly based on the video content, select the single best answer option. You must provide a best answer for this question.

If an ordinal number appears in the provided question, note that, in most cases, its meaning is unrelated to the ordinal position of a provided time period. Focus on analyzing the meaning of the ordinal number within the question.

Please output a strictly standardized dictionary containing exactly three key--value pairs:

\begin{itemize}
    \item \texttt{"Answer"}: A string enclosed in double quotation marks. This must contain exactly one number. If answer options are provided, return the number of the selected option. If no answer options are provided, return the numeric answer to the question.

    \item \texttt{"Time Period"}: A list containing the time periods corresponding to the best answer, with each period represented as a tuple in the format \texttt{(start time, end time)}.

    \item \texttt{"Reason"}: A string enclosed in double quotation marks. Explain the reasoning behind your judgment. Ensure that the reasoning provides complete and objective support for your answer.
\end{itemize}

Do not add any additional comments within the dictionary.

Return \textbf{only} a valid JSON object.

The response must start with \texttt{\{} and end with \texttt{\}}.

Do not wrap the JSON in Markdown fences.

Do not add explanations before or after the JSON.

Use double quotation marks for all keys and string values.

Use JSON Boolean values \texttt{true}/\texttt{false}, not Python Boolean values \texttt{True}/\texttt{False}.

For \texttt{"Time Period"}, use JSON arrays such as \texttt{[[120, 150]]}, not tuples such as \texttt{[(120, 150)]} or strings such as \texttt{"120s-150s"}.
\end{quote}

We preserve VideoLucy’s forced-answer policy while adapting it for cross-dataset evaluation: questions are framed as academic benchmark tasks, answer choices are embedded directly in the question, predictions are standardized as numeric outputs, and responses are constrained to valid JSON for reliable Gemini API parsing.

\subsubsection{Video Curious Agent}

The code for VCA is not public as of August 2026. We reimplemented the method as described in the paper. For prompts we follow the templates from the appendix of ~\citet{yang2025vca}, however we encounter a few inconsistencies in the paper itself as detailed in the exploration prompt below:

\paragraph{Reward Model Prompt for first round}

\begin{quote}
You are acting as a reward model to guide the video question-answering process, with access to a \texttt{\{self.total\_frames\}}-frame video (\texttt{\{self.video\_duration\_sec\}} seconds). You are provided with \(N\) uniformly sampled frames from the video, at the following frame indices: \texttt{\{","{.}join(map(str, frame\_indices))\}}, which divide the video into \(N+1\) distinct segments.

\textbf{Transcript aligned with these sampled frames:}

\texttt{\{transcript\}}

\texttt{\{segment\_text\}}

Your task is to evaluate the relevance of each segment in answering the question below, to assist in identifying the segment(s) that most effectively answer the question.

\texttt{\{question\}}

First provide an explanation based on the specific details observed in each sub-segment, then assign a relevance score from 0 to 100. Focus on each segment‘s alignment with the question.

Hint: Since only the start and end frames of each segment are provided, first imagine the possible content within each segment based on these frames before making a decision.

In your explanation, describe your reasoning process and any inferred details about the segment‘s content, using specific observations from the start and end frames as a basis for these inferences. Please give the answer in the format:
\begin{verbatim}
{"Segment #": {explanation: str, score: int}}
\end{verbatim}
It needs to be a valid Python literal.
\end{quote}

We preserve VCA’s original tree-search, reward-guided exploration, and memory-management mechanisms. To ensure a controlled comparison, we additionally provide timestamp-aligned transcripts to VCA, as done for all evaluated methods. Prompts retain the original
instructions, but explicitly require a standardized Python-literal output format to address inconsistencies observed in preliminary Minerva evaluations.

\paragraph{Reward Model Prompt for subsequent rounds}

\begin{quote}
You are acting as a reward model in a multi-round video question-answering process. You have access to a \texttt{\{self.total\_frames\}}-frame video (\texttt{\{self.video\_duration\_sec\}} seconds), along with results from a previous round of evaluation. In this round, one specific segment has been further divided to provide more detailed analysis. You are provided with \(N\) new sampled frames to assess these sub-segments in relation to the question, at the following frame indices:

\texttt{\{","{.}join(map(str, frame\_indices))\}}

Question:

\texttt{\{question\}}

In the last round, the video was divided into multiple segments, and each segment was evaluated for its relevance to the goal question. Here are the results from all previous rounds:

\texttt{\{reward\_history\_text\}}

In this round, segment [\texttt{\{frame\_indices[0]\}}, \texttt{\{frame\_indices[-1]\}}] has been further explored with new uniformly sampled frames, dividing it into new sub-segments:

\texttt{\{segment\_text\}}

Your task is to evaluate these new sub-segments for relevance to the original goal question based on provided frames, to assist in identifying the segment(s) that most effectively answer the question, while considering the context and results from the previous rounds.

First provide an explanation based on the specific details observed in each sub-segment, then assign a relevance score from 0 to 100. Focus on each segment‘s alignment with the question, ensuring consistency with the scores and explanations from previous segments.

Transcript aligned with these sampled frames:

\texttt{\{transcript\}}

Hint: Since only the start and end frames of each segment are provided, first imagine the possible content within each segment based on these frames before making a decision.

In your explanation, describe your reasoning process and any inferred details about the segment‘s content, using specific observations from the start and end frames as a basis for these inferences. Please give the answer in the format:
\begin{verbatim}
{Segment #: {explanation: str, score: int}}
\end{verbatim}
It needs to be a valid Python literal.
\end{quote}

Again our prompt is almost identical to the original, we only require a stricter output format for consistency and add the transcript for a controlled evaluation.

\paragraph{Exploration Agent Prompt}

\begin{quote}
You are a helpful assistant with access to a video that is \texttt{\{self.total\_frames\}} frames long (\texttt{\{self.video\_duration\_sec\}} seconds).

\texttt{\{question\}}

You are tasked with exploring the video to gather the information needed to answer a specific question with complete confidence. At each step, you may select one segment of the video to examine. Once you choose a segment, you will receive a set of representative frames sampled from that segment. Use each exploration step strategically to uncover key details, progressively refining your understanding of the video's content. Continue exploring as needed until you have acquired all the information necessary to answer the question.

Transcript aligned with current memory frames:

\texttt{\{transcript\}}

Memory buffer frame indices:

\texttt{[\{memory\_frame\_indices\_text\}]}

Candidate segments to choose from:

\texttt{\{candidate\_segments\_text\}}

For each candidate segment, an auxiliary video assistant has already evaluated the relevance score with respect to the question. Focus on the most promising segment if more exploration is needed.

If you need more exploration, respond with a valid Python literal in this format:

\begin{verbatim}
{"action": "explore", "start_idx": int, "end_idx": int}
\end{verbatim}

If you already have enough information, respond with a valid Python literal in this format:

\begin{verbatim}
{"action": "answer", "answer": int, "reasoning": str}
\end{verbatim}

for multiple choice questions, where the answer is the index of the correct choice.

\begin{verbatim}
{"action": "answer", "answer": float, "reasoning": str}
\end{verbatim}

for questions requiring a numerical answer.

Hint: Do not rush to provide an answer. Take time to verify details and gather sufficient information before concluding. Approach each question as a step toward building a comprehensive understanding, ensuring accuracy over speed.

You must always return a valid Python literal.
\end{quote}

This prompt is based on Figure 8 of the original paper~\cite{yang2025vca}. We identify an ambiguity between the presented prompt and the algorithmic description: Figure 8 displays only the subsegments created during the current round, whereas Section 3.3 and Algorithm 1 specify that the exploration agent receives the accumulated candidate set (S), comprising both newly created fine-grained segments and unexplored segments from previous rounds. The latter interpretation is necessary to support the backtracking behavior described in the paper. We therefore follow Algorithm 1 and provide the accumulated candidate set. The remaining instructions follow the original template, apart from the addition of timestamp-aligned transcripts and a standardized output format for reliable parsing.

\subsubsection{VideoAgent}

For the VideoAgent implementation we use chat functionality from Gemini (except for captioning), where for each question the model retains access to all prior  messages sent during the question answering process.

\paragraph{Answering Prompt (Initial Prompt)}

\begin{quote} You will be given a question about a video and some possible answer options. You are provided descriptions of frames sampled from the video at relevant points. Frame Descriptions: \texttt{\{captions\_text\}} Transcript: \texttt{\{transcript\}} Question: \texttt{\{question\}} After explaining your reasoning, output the final answer in the format ``Final Answer: (X)'' where X is the correct digit choice. Never say ``unknown'' or ``unsure'', or ``None'', instead provide your most likely guess. \end{quote}

The caption prompt from the original repository~\footnote{https://github.com/wxh1996/VideoAgent} includes EgoSchema-specific hints such as "\#C" for the camera wearer and \#O for someone other than the camera wearer. For the cross-dataset evaluation we remove those hints, while keeping the semantics of the prompt intact. In addition, as with all other baselines, we add the ASR-generated transcript to the prompt. We also adjust the output format to our evaluation. Preliminary Minerva runs revealed that VideoAgent tends to return "unknown" frequently; to ensure a fair evaluation with other baselines (which are forced to provide a guess), we prohibit VideoAgent from returning an invalid answer .

\paragraph{Self-Evaluation Prompt}

\begin{quote} Please assess the confidence level in your decision-making process for the answer you just provided: ``\texttt{\{answer\}}'' Criteria for Evaluation: Insufficient Information (Confidence Level: 1): If information is too lacking for a reasonable conclusion. Partial Information (Confidence Level: 2): If information partially supports an informed guess. Sufficient Information (Confidence Level: 3): If information fully supports a well-informed decision. Assessment Focus: Evaluate based on the relevance, completeness, and clarity of the provided information in relation to the decision-making context. Return only valid JSON in the format \texttt{\{confidence\_format\}}. The confidence value must be the integer 1, 2, or 3, not a label or explanation. \end{quote}

Apart from a stricter output format, this prompt is identical to the original one from the repository~\footnote{https://github.com/wxh1996/VideoAgent}.

\paragraph{Captioning Prompt}

\begin{quote}
You are an expert video analyst. Describe this video frame in detail, specifically focusing on information relevant to the following context:

\texttt{\{condition\}}

In your description, focus on:
\begin{itemize}
    \item Main subjects/objects visible
    \item Actions or activities occurring
    \item Scene/environment description
    \item Any text visible
    \item Relevant spatial relationships
\end{itemize}

Provide a concise but comprehensive description in 2-3 sentences.
\end{quote}

The original repository and paper~\cite{VideoAgent} do not provide the prompt used to generate captions; instead, they release only precomputed captions for EgoSchema. We therefore introduce the task-conditioned captioning prompt above to generate frame descriptions in our implementation.

\paragraph{Segment Generation Prompt}

\begin{quote} So the information in our previous frames was not sufficient. \textbf{Objective:} Our goal is to identify additional frames that contain crucial information necessary for answering the question. These frames should not only address the query directly but should also complement the insights from the previous frames. To achieve this, we will: 1. Divide the video into candidate segments based on the intervals between the initial frames. Candidate Segments (ID: Start-End in Indices): \texttt{\{segment\_des\}} 2. Determine which segments are likely to contain frames that are most relevant to the question. These frames should capture key visual elements (objects, humans, interactions, actions, scenes) that help answer the question. \textbf{Instructions:} \begin{itemize} \item For each frame identified as potentially relevant, provide a concise description focusing on essential visual elements. \item Use a single sentence per frame. \item If the specifics of a segment's visual content are uncertain based on the current information, use placeholders for specific actions or objects, but ensure the description still conveys the segment's relevance to the query. \item You may select multiple frames from one segment if necessary. \end{itemize} Return the descriptions and the \texttt{segment\_id} in JSON format. \begin{itemize} \item \texttt{"segment\_id"} must be an integer from the Candidate Segments keys (1 to \texttt{\{len(segment\_des)\}}). \item \texttt{"duration"} must match the duration string from the chosen candidate segment. \end{itemize} \textbf{Example Output:} \begin{verbatim} { "frame_descriptions": [ 
{ "segment_id": 1, "duration": "xxx - xxx", 
"description": "frame of xxx" }, 
{ "segment_id": 2, "duration": "xxx - xxx", 
"description": "frame of xxx" }, 
{ "segment_id": 3, "duration": "xxx - xxx", 
"description": "frame of xxx" } 
] } \end{verbatim} Output just the JSON object, nothing else. \end{quote}

Semantically, this prompt is equivalent to the original. The caption and question are retained from previous prompts through the chat functionality.

\subsubsection{Gemini Native}

\paragraph{Multiple-choice prompt}

\begin{quote} You will be given a question about a video and five possible answer options. Watch the video carefully to answer the question. Question: \texttt{\{question\}} After explaining your reasoning, output the final answer in the format ``Final Answer: (X)'' where X is the correct digit option choice. Never say ``unknown'' or ``unsure'', or ``None'', instead provide your most likely guess. \end{quote}

\paragraph{Numeric questions prompt (VSIBench)}

\begin{quote} You are an expert at spatial reasoning. Watch the video carefully and answer the following numerical question. Think step by step before giving your final answer. \texttt{\{question\}} Output your final answer as a number in the format ``Final Answer: <number>'' (e.g., ``Final Answer: 42.5''). \end{quote}

Both prompts are adapted from the original Minerva paper~\cite{nagrani2025minerva}.

\subsubsection{DeepVideoDiscovery}

To adapt DeepVideoDiscovery to the Gemini API we use the Gemini analogue of OpenAI tools, \texttt{types.Tool}.

\paragraph{Captioning Prompt}

\begin{quote}
There are consecutive frames from a video. Please understand the video clip with the given transcript then output JSON in the template below.

\textbf{Transcript of current clip:}

\texttt{TRANSCRIPT\_PLACEHOLDER}

\textbf{Output template:}

\begin{verbatim}
{
  "clip_start_time": "CLIP_START_TIME",
  "clip_end_time": "CLIP_END_TIME",
  "subject_registry": {
    <subject_i>: {
      "name": <fill with short identity if name is unknown>,
      "appearance": <list of appearance descriptions>,
      "identity": <list of identity descriptions>,
      "first_seen": <timestamp>
    },
    ...
  },
  "clip_description": <smooth and detailed natural narration 
  of the video clip>
}
\end{verbatim}

Remember your output MUST be a valid JSON.
\end{quote}

We use a near-identical copy of the prompt from the original repository~\footnote{\url{https://github.com/microsoft/DeepVideoDiscovery/tree/64414b2f35d26809a39740a5a319889f46e29b94}}, with only two formatting-related modifications: timestamp placeholders are quoted, and the model is
explicitly instructed to produce valid JSON. We observed that this significantly increased output reliability, necessary for parsing model responses.

\paragraph{Subject-registry merge prompt}

\begin{quote}
You are given several partial \texttt{new\_subject\_registry} JSON objects extracted from different clips of the \emph{same} video. They may contain duplicated subjects with slightly different IDs or descriptions.

\textbf{Task:}

\begin{enumerate}
    \item Merge these partial registries into one coherent \texttt{subject\_registry}.
    \item Preserve all unique subjects.
    \item If two subjects obviously refer to the same person, merge them
    (keep earliest \texttt{first\_seen} time and union all fields).
\end{enumerate}

The merge prompt is an exact copy of the original prompt.

\paragraph{Main Agent Prompt}

\begin{quote}
You are a helpful assistant who answers multi-step questions by sequentially invoking the provided function tools. Follow the THINK $\rightarrow$ ACT $\rightarrow$ OBSERVE loop:

\begin{itemize}
    \item \textbf{THOUGHT} Reason step-by-step about which function to call next.
    \item \textbf{ACTION} Invoke exactly one provided function tool that moves you closer to the final answer.
    \item \textbf{OBSERVATION} Summarize the function's output.
\end{itemize}

Use native function calling only. Do not write code, \texttt{print(...)}, \texttt{default\_api} calls, or textual ACTION calls.

You MUST plan extensively before each function call, and reflect extensively on the outcomes of the previous function calls.

Only pass arguments that come verbatim from the user or from earlier function outputs---never invent them. Continue the loop until the user's query is fully resolved, then end your turn with the final answer. If you are uncertain about code structure or video content, use the available tools to inspect rather than guessing. Plan carefully before each call and reflect on every result. Do not rely solely on blind function calls, as this degrades reasoning quality. Timestamps may be formatted as \texttt{'HH:MM:SS'} or \texttt{'MM:SS'}.

Carefully read the timestamps and narration in the following script, paying attention to the causal order of events, object details and movements, and people's actions and poses.

Here are tools you can use to reveal your reasoning process whenever the provided information is insufficient.

\begin{itemize}
    \item To get a global information about events and main subjects in the video, use \texttt{global\_browse\_tool}.
    \item To search without a specific timestamp, use \texttt{clip\_search\_tool}.
    \item If the retrieved material lacks precise, question-relevant detail (e.g., an unknown name), call \texttt{frame\_inspect\_tool} with a list of time ranges (\texttt{list[tuple[HH:MM:SS, HH:MM:SS]]}).
    \item Whenever you are uncertain of an answer after searching, inspect frames in the relevant intervals with \texttt{frame\_inspect\_tool}.
    \item After locating an answer in the script, always make a \textbf{CONFIRM} with \texttt{frame\_inspect\_tool} query.
\end{itemize}

You can first use \texttt{global\_browse\_tool} to a global information about this video, then invoke multiple times of these tools to progressively find the answer.

Based on your observations and tool outputs, provide a concise answer that directly addresses the question.

ALWAYS invoke one of the provided function tools. Do not write tool calls as text. If you're done, invoke the \texttt{'finish'} tool.

After either you've called the \texttt{'finish'} tool or you've reached the max number of iterations, you'll be prompted to answer the question. Here please DON'T call a tool just provide your answer in text.

Total video length: \texttt{VIDEO\_LENGTH} seconds.

Question: \texttt{QUESTION\_PLACEHOLDER}
\end{quote}
]

\textbf{Input (list of JSON objects):}

\texttt{REGISTRIES\_PLACEHOLDER}

Return \emph{only} the merged \texttt{subject\_registry} JSON object.
\end{quote}

We adjust the wording slightly to use Gemini terminology (\texttt{tools} instead of \texttt{functions}). Furthermore, we encountered several cases in which the model produced textual tool calls or answered prematurely instead of using Gemini's native function-calling interface. We therefore
explicitly require native tool calls and instruct the model to invoke the \texttt{finish} tool before producing its final answer. Apart from these
interface-specific adjustments, the original reasoning and tool-use instructions remain unchanged.

\paragraph{Global-browse Prompt}

\begin{quote}
You are a knowledgeable assistant specializing in analyzing video content and providing detailed, insightful answers.

Below are descriptions of video clips, each with its corresponding timestamp. Carefully review the sequence of events, the details and movements of objects, and the actions and poses of people. Based on these observations, provide a thorough and specific answer to the following question, referencing key events and timestamps as appropriate.

Question: \texttt{\{query\}}

\texttt{\{clip\_captions\}}
\end{quote}

No changes were made to this prompt.

\paragraph{Frame-inspection Prompt}

\begin{quote}
Carefully watch the video frames. Pay attention to the cause and sequence of events, the detail and movement of objects and the action and pose of persons.

Based on your observations, if you find content that can answer the question. If no relevant content is found within the given time range, return:

\texttt{Error: Cannot find corresponding result in the given time range.}

Question: \texttt{\{question\}}
\end{quote}

Again, we do not modify the original prompt.

\subsubsection{A.I.R.}

We copy the algorithm from the authors' official GitHub~\footnote{https://github.com/UCF-AIR/A.I.R./tree/main} repository adapting only the backbone calls as required by our evaluation protocol.

\paragraph{Analysis Prompt}

\begin{quote}
You are an expert visual reasoner. Step by step analyze how well the provided images match the following query (with options):

Query: \texttt{\{query\}}

Rate the relevance from 1 to 5, using these exact definitions:

1 – Not relevant at all: no relation between image content and the query.

2 – Slightly relevant: only minor contextual hints, but not central to answering.

3 – Moderately relevant: contains preparatory or follow-up context (e.g. setup actions) related to the query.

4 – Highly relevant: shows clear evidence to support query but still has ambiguity or missing detail.

5 – Perfectly matches: fully sufficient and unambiguous evidence to answer the query.

Avoid overthinking. Trust your immediate judgment.

Examples:

– Image of an empty room (no people or objects) → Reasoning: There is nothing related to the query. Score: 1

– Image showing exactly the queried action (clear, direct match) → Reasoning: It perfectly depicts the required event. Score: 5

Respond with exactly one valid JSON array and nothing else.

Requirements:
\begin{itemize}
    \item Return exactly one object for every provided frame.
    \item Use the exact zero-based Index shown in each frame header.
    \item Score must be an integer from 1 through 5.
    \item Do not include text before or after the JSON array.
\end{itemize}

Example:
\begin{verbatim}
[
    {
        "Index": 0,
        "Score": 4,
        "Reasoning": "The frame clearly shows evidence 
        relevant to the question."
    }
]
\end{verbatim}
\end{quote}

The analysis prompt is copied verbatim from the repository, only the output formatting is adjusted to ensure robustness to common parsing and consistency failures encountered during evaluation. We specifically require the model to output its analysis in JSON format, consistent with requirements for other baselines.

\section{Application of the Active Exploration Formalism to Existing Approaches}
\label{sec:formalism_application}
\subsection{Active Exploration Baseline}
\label{sec:appendix_active_exploration_baseline}
We subsequently present one of the simplest formalizations of the agentic video reasoning framework which we use to construct the Active Exploration Baseline.
\paragraph{Action Space and Observations}
The action space consists of the simplest conceivable operator:
\[
\mathcal{A}_{\text{query}} = \{\textsc{Query}(n)\}
\]

\(\textsc{Query}\) takes in a frame index \(n\) and simply returns the corresponding frame from the video.
The observation operator \(O\) therefore returns the frame at index \(n\):
\[
o_{t+1} = O(\textsc{Query}(n), E) = x_n
\]
where \(x_n\) is the frame at index \(n\).

\paragraph{Initialization}
We don't use extensive video preprocessing, \(F_{\text{init}}\) outputs the first frame index to inspect, along with the agent's reasoning \(r_1\) (without having seen any frames):

\[
s_1 =  F_{\text{init}}(q, E) = (n_1, r_1)
\qquad
\pi(\textsc{Query}(n_1)\mid s_1)=1.
\]

\paragraph{State Update}
The state update is the primary exploration step. The agent chooses the next action and outputs its internal reasoning, both of which are concatenated with all prior reasoning outputs, giving the interaction history up to iteration \(t+1\), \(r_{t+1}\).

\[
s_{t+1} = F(q, s_t, o_{t+1}) = (a_{t+1}, r_{t+1})
\]

\(a_{t+1}\) is the next action, either \(\textsc{Stop}\) or \(\textsc{Query}(n_{t+1})\).

\paragraph{Policy and termination}

The policy then deterministically extracts \(a_{t+1}\) from \(s_{t+1}\). 
Upon termination the agent executes a final reasoning step based on the interaction history and outputs a final answer. \(\tau\) denotes termination time:

\[
\hat{y} = g(s_{\tau}) \in \mathcal{Y}.
\]

\subsection{VideoAgent}
\paragraph{Action Space and Observations}
In VideoAgent \cite{VideoAgent} the actions are presented as:
\[
\mathcal{A}_{\text{query}} = \{\textsc{UniformSampling}, \textsc{SegmentSearch}(Q)\}
\]

Here, \textsc{UniformSampling} returns uniformly sampled frames from the full video, while
\textsc{SegmentSearch} takes in a set of queries \(Q = (q_i, seg_i)\). For each query \((q_i, seg_i)\), where \(q_i\) denotes a textual query and \(seg_i\) specifies a temporal segment in the video, the action retrieves the frame within the corresponding segment, which is most similar according to CLIP \cite{radford2021learning} similarity. The observation operator \(O\) returns the captions of the returned frames for either action.

\paragraph{Initialization}
Since VideoAgent does not include a dedicated preprocessing step, we model initialization as returning a fixed start state. Concretely, we set

\[
s_1 = F_{\text{init}}(q, E) = s^{(0)}
\]

where \(s^{(0)}\) is a special \textsc{Start} token such that the policy selects \textsc{UniformSampling} with probability one:

\[
\pi(\textsc{UniformSampling} \mid s^{(0)}) = 1
\]

\paragraph{State Update}
The state update \(F\) is VideoAgent's internal reasoning step: At time \(t\) it tries to answer the question and assess its own confidence. If the confidence surpasses a certain threshold \textsc{Stop} is invoked and the final answer is returned in \(s_{t}\), otherwise the process continues with \textsc{SegmentSearch}. In either case the new captions are added to the internal memory from \(s_{\text{t-1}}\). \(s_t\) consists of the frame captions, the answer to the user's question and the agent's confidence in the answer.

VideoAgent's policy is deterministic:
\[
\text{Either} \qquad \pi(\textsc{Stop}\mid s_t)=1
\quad \text{or} \quad
\pi(\textsc{SegmentSearch}\mid s_t)=1.
\]
depending on which action the agent chose in \(F\). Note that VideoAgent uses \textsc{UniformSampling} only for initialization; our framework also permits a straightforward extension where \textsc{UniformSampling} becomes a reusable query action.

\paragraph{Final prediction}
The prediction head just extracts the correct answer from the final agent state \(s_{\tau}\)

\[
\hat{y} = g(s_{\tau})
\]

\subsection{Video Curious Agent}

VCA~\cite{yang2025vca} treats a long video as an environment to explore via a tree-search over temporal segments.
At each iteration it uniformly samples frames from the currently selected segment, scores the resulting subsegments using an intrinsic reward model, updates a fixed-size memory buffer, and then either outputs an answer or selects a new segment for further exploration.

\paragraph{Action space.}
In our formalism, VCA's query actions correspond to selecting the next temporal segment to explore.
Let a segment be parameterized by its temporal bounds \((\ell,r)\). We write
\[
\mathcal{A}_{\text{query}} = \{\textsc{SelectSegment}(\ell, r)\},
\qquad
\mathcal{A}=\mathcal{A}_{\text{query}}\cup\{\textsc{Stop}\}.
\]

\paragraph{Observation operator.}
Given the current segment \((\ell, r)\), VCA observes the environment by sampling \(N\) frames uniformly from that segment ~\cite{yang2025vca}. We model this as
\[
o_{t+1}=O(\textsc{SelectSegment}(\ell_t, r_t),E)=\textsc{UniformSample}(\ell_t, r_t),
\]
where \(o_{t+1}\in\mathcal{O}\) contains the sampled frames.

\paragraph{Initialization.}
VCA initializes the agent state to contain uniformly sampled frames across the whole video. We capture this by an initializer
\[
s_1 = F_{\text{init}}(q, E) = s^{(0)},
\]
where \(s^{(0)}\) is a special \textsc{Start} token such that the policy selects \textsc{SelectSegment} on the whole video with probability one:

\[
\pi(\textsc{SelectSegment}, \ell_0 = 0, r_0=N \mid s^{(0)}) = 1
\]

where \(N\) is the last frame of the video.

\paragraph{State update.}
VCA's state update corresponds to (i) computing intrinsic reward scores for candidate subsegments using a reward model \(R\), (ii) updating the candidate set and reward-history (lines 12--14), and (iii) updating a fixed-size memory buffer with sampled frames (line 16)~\cite{yang2025vca}. In our notation,
\[
s_{t+1}=F(q,s_t,o_{t+1}),
\]
where \(s_t\) can be understood as the agent's state, containing candidate segments, the reward history, the fixed-size memory buffer, and the agent's answer to the question.

\paragraph{Policy and termination.}
The induced policy selects either \(\textsc{Stop}\) or \(\textsc{SelectSegment}(\cdot)\); termination occurs when a non-empty answer is present in \(s_{t+1}\).

\paragraph{Prediction head.}
Upon termination at time \(\tau\), the agent has returned a non-empty answer which is extracted from \(s_{\tau}\):

\[
\hat{y}=g(s_{\tau}).
\]

\subsection{Deep Video Discovery (DVD)}
Deep Video Discovery (DVD)~\cite{zhang2025deep} frames long-video understanding as iterative
information gathering over a multi-granular video database using a search-centric toolset and an LLM
orchestrator. Concretely, the agent iteratively selects an action from a toolset (or answers) and
updates its internal reasoning based on the returned tool outputs.

\paragraph{Action space}
DVD exposes three query tools—\textsc{GlobalBrowse}, \textsc{ClipSearch}, and \textsc{FrameInspect}, in addition to the \textsc{Stop} action from our formalism.
We therefore model

\begin{align*}
&\mathcal{A}_{\text{query}} =
\{\textsc{GlobalBrowse}(s_t,q),\;
  \textsc{ClipSearch}(s_t,\hat{q}),\;
  \textsc{FrameInspect}(s_t,\hat{q}, I)\}
\\
&\mathcal{A}=\mathcal{A}_{\text{query}}\cup\{\textsc{Stop}\}.
\end{align*}

Here \(s_t\in\mathcal{S}\) denotes the current agent state (memory / context),
\(q\in\mathcal{Q}\) is the original user query, and \(\hat{q}\) is a refined sub-query produced by the agent (e.g., a more specific retrieval prompt).
Finally, \(I\) denotes a temporal interval to inspect, e.g. \(I=[t_s,t_e]\).
Intuitively, \textsc{GlobalBrowse} produces a coarse global summary conditioned on \((s_t,q)\),
\textsc{ClipSearch} retrieves relevant clips for \(\hat{q}\), and \textsc{FrameInspect} performs a targeted inspection within \(I\) for \(\hat{q}\).

\paragraph{Observation operator}
The observation operator \(O\) returns the tool outputs:
(i) \textsc{GlobalBrowse} returns global summaries (subject-centric and event-centric),
(ii) \textsc{ClipSearch} returns top-$k$ relevant clips along with captions and time ranges via caption-embedding
retrieval, and
(iii) \textsc{FrameInspect} returns an open-format VQA response for a specified time window and sub-query.
Formally, for a tool action \(a_t\in\mathcal{A}_{\text{query}}\),
\[
o_{t+1}=O(a_t,E)\in\mathcal{O}.
\]

\paragraph{Initialization}
For DVD, the agent first constructs a multi-granular video database that is subsequently queried by the tool actions.
\[
s_1 = F_{\text{init}}(q, E).
\]

Importantly, this preprocessing step is query-agnostic, so \(F_{\text{init}}\) discards the input \(q\). The resulting state \(s_1\) stores the multi-granular video representations that are used by the subsequent tool calls.

\paragraph{State update}
The state update \(F\) corresponds to the LLM's iterative reasoning and memory update: it integrates the
new observation \(o_{t+1}\) (tool output) into the current reasoning context and decides what information is
still missing:
\[
s_{t+1} = F(q,s_t,o_{t+1})\in\mathcal{S}.
\]

The updated state \(s_{t+1}\) contains the database as well as the interaction history, consisting of reasoning traces, actions, and observations. Consequently, the size of \(s_{t+1}\) grows linearly with the number of interaction steps.

\paragraph{Policy and termination}
DVD's policy \(\pi\) is deterministic and directly extracts the action from the previous state.

\paragraph{Final prediction}
Upon termination at time \(\tau\), the final answer is extracted from the terminal state \(s_\tau\).

\subsection{VideoTree}

VideoTree~\cite{wang2025videotree} constructs a query-adaptive hierarchical representation of the video by
first building a set of query-relevant top-level clusters and then expanding only the relevant clusters
to finer levels before performing final LLM reasoning.

\paragraph{Action space}
In our formalism, VideoTree's query actions correspond to operations that iteratively refine a
tree-structured representation of the video. Instead of directly selecting frames or segments,
the agent acts by modifying this tree.

We introduce four key variables: \(k\), the number of clusters used at the current breadth
expansion step; \(C\), a cluster (i.e., a node in the tree); \(w\), the branching factor; and \(d\), the depth of expansion. Based on this, we define the query action space as follows:
\begin{align*}
& \mathcal{A}_{\text{query}}=
\{
\textsc{BreadthExpand}(k),\;
\textsc{DepthExpand}(C,w,d),\;
\textsc{CaptionKeyframes}
\},
\\
& \mathcal{A}=\mathcal{A}_{\text{query}}\cup\{\textsc{Stop}\}.
\end{align*}

Here, \textsc{BreadthExpand}(k) performs visual clustering of the video into \(k\) clusters and returns the caption of each keyframe; \textsc{DepthExpand}(C,w,d) refines a selected cluster \(C\) into a subtree
of width \(w\) and depth \(d\); and \textsc{CaptionKeyframes} traverses the constructed tree, extracts representative keyframes from its clusters,
and returns their captions in temporal order.

\paragraph{Observation operator}
The observation operator returns the outputs of these refinement steps:
\[
o_{t+1}=O(a_t,E)\in\mathcal{O}.
\]
Concretely:
(i) \textsc{BreadthExpand}(k) returns top-level clusters together with representative captions;
(ii) \textsc{DepthExpand}(C,w,d) returns a refined subtree of depth \(d\) rooted at \(C\), obtained by iteratively re-clustering the
frames in \(C\);
and (iii) \textsc{CaptionKeyframes} returns captions for the keyframes in temporal order.

Thus, \(\mathcal{O}\) contains structured observations such as clusters, captions and
subtrees, rather than only raw frames.

\paragraph{Initialization}
VideoTree initializes the agent state with a fixed start token. Concretely, we set
\[
s_1 = F_{\text{init}}(q,E) = s^{(0)},
\]
where \(s^{(0)}\) is a special \textsc{Start} token such that the policy selects \textsc{BreadthExpand} for the initial clustering with probability one:
\[
\pi(\textsc{BreadthExpand}(k_0) \mid s^{(0)}) = 1.
\]

\paragraph{State update}
The state update
\[
s_{t+1}=F(q,s_t,o_{t+1})
\]
corresponds to updating the current tree representation and its associated metadata.
The state \(s_t\) contains the partially constructed VideoTree, including current clusters,
their relevance scores, representative keyframes and captions.
During adaptive breadth expansion, \(F\) updates the first-level clustering by providing the relevance scores between 1 and 3 for each cluster.
If the observation \(o_{t+1}\) contains the output of \textsc{CaptionKeyframes}, the state update $F$ stores these captions in a temporally ordered reasoning context within $s_{t+1}$ and prepares the agent for termination.

\paragraph{Policy and termination}
VideoTree's induced policy is deterministic.
If the state update \(F\) indicates that the number of highly relevant clusters (relevance score 3) is below a threshold and \(k\) is below the maximum breadth, the policy chooses \textsc{BreadthExpand}(k).
Otherwise, it selects \textsc{DepthExpand}(C,w,d) for clusters based on the relevance scores assigned during the state update: \(d=1\) for score 2 and \(d=2\) for score 3.
Once the tree refinement is complete, the policy selects \textsc{CaptionKeyframes}; upon receiving the final captions, it chooses \textsc{Stop}.

\paragraph{Final prediction}
Upon termination at time \(\tau\), the final answer is produced from the terminal state:
\[
\hat{y}=g(s_\tau).
\]
Here, \(g\) corresponds to the LLM reasoning module that consumes the temporally ordered captions
extracted from the constructed tree together with the original query and outputs the answer.

\subsection{VideoLucy}

VideoLucy~\cite{zuo2025videolucy} addresses long-video question answering through a deep backtracking mechanism. It builds a coarse video memory by captioning the full video at a low temporal resolution, and then iteratively re-examines the most relevant temporal periods at
progressively finer granularity until a confident answer emerges.                                               

\paragraph{Action space}

At each step, VideoLucy selects a temporal period to re-examine together with a targeted captioning
instruction. We define
\begin{align*}
&\mathcal{A}_{\text{query}} = \{\textsc{RecaptionPeriod}(p,I) \mid p=(s,e)\subseteq[0,T],\;I\in\mathcal{I}\}, \\
&\mathcal{A} = \mathcal{A}_{\text{query}}\cup\{\textsc{Stop}\}.
\end{align*}
Here $p=(s,e)$ is a temporal interval and $I$ is a natural-language captioning instruction specifying which visual details to focus on.

  \paragraph{Observation operator}
  Given action $\textsc{RecaptionPeriod}(p,I)$, the observation operator produces hierarchical captions
  of period $p$:
  \[
  o_{t+1}=O(\textsc{RecaptionPeriod}(p,I),E)\in\mathcal{O}.
  \]
  Concretely, $o_{t+1}$ consists of (i) an \emph{entire} caption of the full period $p$ at a
  resolution determined by its duration, and (ii) \emph{divided} captions of non-overlapping sub-clips
  of $p$ at a finer temporal resolution.

  \paragraph{Initialization}
  VideoLucy's initialization is a two-phase preprocessing step.

  Specifically $\mathrm{CM}=\{(p_k,d_k)\}_{k=1}^{K_c}$ is first constructed
  where each entry stores a temporal period $p_k$ and its caption $d_k$.
  In $F_{\text{init}}$, the agent tries to answer the question based solely on $\mathrm{CM}$. Two cases are distinguished: (1) If the agent can confidently answer the question just based on $\mathrm{CM}$, $F_{\text{init}}$ returns a special termination state $s^{(0)}$.
  (2) Otherwise, $\mathrm{CM}$ is internally evaluated against the query $q$ to extract the top-$K$ most relevant periods, forming a sparse coarse memory $\mathrm{CM}_{\text{init}}\subseteq\mathrm{CM}$ which populates the standard initial state $s_1$ as the working memory for the subsequent exploration loop. We can therefore write:
  
  \[
  s_1 = F_{\text{init}}(q, E).
  \]

  \paragraph{State update}
  At each iteration, the state update $F$ integrates the newly acquired fine-grained observation
  $o_{t+1}$ into a three-level hierarchical memory. Then, an Answering Agent evaluates whether this updated memory suffices to answer the question confidently. Consequently, the new state
  \[
  s_{t+1}=F(q,s_t,o_{t+1}).
  \]
  stores the updated hierarchical memory (combining $\mathrm{CM}_{\text{init}}$ with fine and super-fine histories), the drafted answer, and the Answering Agent's confidence score.

  \paragraph{Policy and termination}
  If the initialization returns the special state $s^{(0)}$, the policy immediately terminates with $\pi(\textsc{Stop}\mid s^{(0)})=1$.
  Otherwise, for a working state $s_{t+1}$, the policy's distribution is determined by the recorded confidence. If the Answering Agent was confident, the system terminates. Otherwise, a Localization+Instruction Agent processes $s_{t+1}$ to select the single most relevant unexplored period and produce a targeted captioning instruction, determining the next $\textsc{RecaptionPeriod}(p^*,I^*)$ action:
  \begin{align*}
  &\pi(\textsc{Stop}\mid s_{t+1})=1 \quad\text{if } s_{t+1} \text{ is confident, else} \\
  &\pi(\textsc{RecaptionPeriod}(p^*,I^*)\mid s_{t+1})=1.
  \end{align*}

  \paragraph{Final prediction}
  Upon termination at time $\tau$, the final answer is extracted from the terminal state:
  \[
  \hat{y}=g(s_\tau).
  \]

\section{Validation Against Prior Baseline Results}
\label{sec:result_validation}

To sanity-check our adaptations of the baselines to the new backbones, we compare our Gemini 3 Flash and Gemma 4 results to publicly reported results of the same baselines using other backbones.
\paragraph{VideoAgent:} Our VideoAgent implementation with Gemini~3~Flash achieves \textbf{64.2\%} accuracy on the 500-question EgoSchema subset. The original VideoAgent paper~\cite{VideoAgent} reports \textbf{60.2\%} using GPT-4, while VCA reports \textbf{63.2\%} with a GPT-4o-based reimplementation and \textbf{65.2\%} using Qwen2-VL-72B. Gemini~3~Flash is a substantially newer and stronger multimodal model than the GPT-4 and GPT-4o backbones used in earlier evaluations, as indicated by recent multimodal benchmarks~\cite{yue2024mmmupro}. Similarly, our VideoAgent implementation with Gemma~4, which is a frontier open-weight model, achieves \textbf{67.7\%} on EgoSchema. The modest improvements over the original GPT-4 result and the GPT-4o reimplementation are therefore directionally consistent with the use of newer backbones. At the same time, the differences are small: on the 500-question subset, one percentage point corresponds to only five examples. Furthermore, direct comparability to prior results is limited, as we are only able to evaluate on 491 of the original 500 EgoSchema samples, and the VCA reimplementation is not publicly available. We therefore interpret these results primarily as evidence that our implementation reproduces the expected VideoAgent performance regime.

\paragraph{VideoTree:} Our VideoTree implementation achieves \textbf{73.1\%} accuracy on the EgoSchema subset with Gemini~3~Flash and \textbf{71.1\%} with Gemma~4. These results are consistent with previously reported VideoTree evaluations using different reasoning backbones. The original VideoTree paper~\cite{wang2025videotree} reports \textbf{66.2\%} with GPT-4, while VCA~\cite{yang2025vca} reports \textbf{67.0\%} with a GPT-4o-based reimplementation and \textbf{71.0\%} using Qwen2-VL-72B. Our Gemini~3~Flash and Gemma~4 results are therefore broadly consistent with prior VideoTree reimplementations, with the improvement over GPT-4/GPT-4o being directionally consistent with the use of newer and stronger multimodal backbones.

\paragraph{VideoLucy:} We could not identify prior VideoLucy evaluations on EgoSchema, Minerva, or VSI-Bench. Consequently, no directly comparable published results are available for validating our VideoLucy reimplementation on these benchmarks.

\paragraph{VideoCuriousAgent (VCA):} VCA does not provide an official public implementation, requiring us to reimplement the method from the paper description and prompts. Prior reported EgoSchema results include \textbf{73.6\%} with GPT-4o and \textbf{75.2\%} with Qwen2-VL-72B in the original VCA paper~\cite{yang2025vca}, as well as \textbf{74.1\%} from the independent EEA reimplementation using Seed1.6VL~\cite{yang2025eeaexplorationexploitationagentlong}. Our implementation achieves \textbf{72.3\%} with Gemini~3~Flash and \textbf{67.7\%} with Gemma~4. Notably, Gemini~3~Flash and Seed1.6VL exhibit broadly comparable performance under the controlled SIS-Bench evaluation~\cite{zou2026selfspacebenchmarkingselfawareness}, further validating the consistency of our Gemini~3~Flash result with the independently reported Seed1.6VL reimplementation. Due to the prohibitive computational cost of unrestricted iterative exploration, we impose a maximum number of VCA iterations. Neither the original VCA paper nor the EEA reimplementation reports whether a hard iteration limit was used, making this aspect of their evaluation setup unclear. Moreover, we evaluate on only 491 of the original 500 EgoSchema samples. Consequently, the results are not directly comparable, but remain in the same general performance regime as prior VCA implementations.

\paragraph{DeepVideoDiscovery (DVD):} The original DVD paper reports \textbf{76.6\%} on EgoSchema using GPT-4.1-mini to construct the textual video database and o3 for agentic reasoning and frame inspection. Our implementation replaces these components with Gemini~2.5~Flash-Lite and Gemini~3~Flash, respectively, and achieves \textbf{69.0\%}. This gap is consistent with the strong backbone sensitivity reported by DVD itself: on LVBench, replacing o3 with weaker reasoning models reduces performance by several percentage points. Since our adaptation changes both the database-construction and reasoning backbones, an exact reproduction of the original score is not expected. We therefore view the observed result as consistent with the performance regime reported for DVD under different model configurations.

\paragraph{Gemini~3~Flash:} The Seed2.0 model card~\cite{seed2026seed20modelcardintelligence} reports Gemini~3~Flash native performance on Minerva at 64.4\%, very close to the 64.8\% obtained with our implementation. This close agreement with an independently reported evaluation provides additional validation of our evaluation pipeline.

\section{Full Results}
\label{sec:full_results}
A full breakdown of the results is shown in Table~\ref{tab:all-video-reasoning-stats}.

\begin{table*}[t]
\centering
\scriptsize
\caption{Comparison of performance, token efficiency, and inference cost across video reasoning benchmarks and backbones.}
\label{tab:all-video-reasoning-stats}

\setlength{\tabcolsep}{4pt}
\renewcommand{\arraystretch}{0.95}

\begin{tabular}{lccccc}
\toprule
Model & Performance (\%) &
\shortstack{Input\\Tokens} &
\shortstack{Output\\Tokens} &
\shortstack{Total\\Tokens} &
\shortstack{Inference\\Cost} \\
\midrule

\multicolumn{6}{l}{\textbf{Gemini 3 Flash}} \\

\addlinespace[2pt]
\multicolumn{6}{l}{\emph{Minerva}} \\
Gemini 3 Flash Native~\citet{gemini3_2026}
    & \underline{64.8} & 61.1K & 9.5K & 70.6K & \$80.54 \\
VideoTree~\citet{wang2025videotree}
    & 50.4 & 99.8K & \underline{6.2K} & 106.0K & \$93.39 \\
VideoAgent~\citet{VideoAgent}
    & 38.5 & \textbf{6.7K} & 7.3K & \textbf{14.1K} & \$34.72 \\
VideoLucy~\citet{zuo2025videolucy}
    & 56.5 & 141.1K & 21.6K & 162.7K & \$154.13 \\
VideoCuriousAgent~\citet{yang2025vca}
    & 51.7 & 97.1K & 41.4K & 138.5K & \$235.76 \\
DeepVideoDiscovery~\citet{zhang2025deep}
    & 52.0 & 68.4K & 16.8K & 85.2K & \underline{\$30.47} \\
A.I.R.~\citet{zou2026air}
    & 55.0 & 41.0K & 2.7K & 43.7K & \$38.95 \\
SAOS
    & 57.5 & \underline{15.6K} & \textbf{0.6K} & \underline{16.2K} & \textbf{\$13.12} \\
\midrule
Active Exploration Baseline
    & 52.8 & 29.4K & 4.2K & 33.6K & \$38.08 \\
SAVE
    & \textbf{66.4} & 38.3K & \underline{1.5K} & 39.7K & \underline{\$32.04} \\

\midrule
\midrule

\addlinespace[4pt]
\multicolumn{6}{l}{\emph{EgoSchema}} \\
Gemini 3 Flash Native~\citet{gemini3_2026}
    & \underline{75.4} & 11.7K & 2.0K & 13.7K & \$5.75 \\
VideoTree~\citet{wang2025videotree}
    & 73.1 & 103.4K & 3.9K & 107.3K & \$31.13 \\
VideoAgent~\citet{VideoAgent}
    & 64.2 & \textbf{2.5K} & 2.8K & \textbf{5.3K} & \underline{\$4.76} \\
VideoLucy~\citet{zuo2025videolucy}
    & 72.7 & 112.9K & 16.6K & 129.5K & \$27.39 \\
VideoCuriousAgent~\citet{yang2025vca}
    & 72.3 & 83.9K & 21.1K & 105.0K & \$51.75 \\
DeepVideoDiscovery~\citet{zhang2025deep}
    & 69.0 & 111.4K & 13.0K & 124.4K & \$11.53 \\
A.I.R.~\citet{zou2026air}
    & 73.9 & 39.7K & 1.9K & 41.6K & \$12.81 \\
SAOS
    & 73.1 & 18.8K & \textbf{0.5K} & 19.3K & \$5.46 \\
\midrule
Active Exploration Baseline
    & 70.0 & \underline{9.1K} & 1.1K & \underline{10.2K} & \textbf{\$3.89} \\
SAVE
    & \textbf{76.0} & 21.6K & \underline{0.9K} & 22.5K & \$6.64 \\

\midrule
\midrule

\addlinespace[4pt]
\multicolumn{6}{l}{\emph{VSIBench (tiny)}} \\
Gemini 3 Flash Native~\citet{gemini3_2026}
    & \textbf{59.2} & \textbf{8.7K} & 7.3K & \textbf{16.0K} & \$10.47 \\
VideoTree~\citet{wang2025videotree}
    & 45.5 & 76.7K & 4.0K & 80.7K & \$20.15 \\
VideoAgent~\citet{VideoAgent}
    & 40.6 & \underline{10.1K} & 7.2K & \underline{17.2K} & \$10.62 \\
VideoLucy~\citet{zuo2025videolucy}
    & 48.2 & 117.0K & 12.5K & 129.5K & \$35.37 \\
VideoCuriousAgent~\citet{yang2025vca}
    & 45.6 & 101.6K & 34.1K & 135.6K & \$61.19 \\
DeepVideoDiscovery~\citet{zhang2025deep}
    & 51.9 & 29.3K & 5.1K & 34.5K & \underline{\$5.21} \\
A.I.R.~\citet{zou2026air}
    & 49.9 & 37.1K & 2.1K & 39.2K & \$9.88 \\
SAOS
    & 50.0 & 16.8K & \textbf{0.8K} & 17.6K   & \textbf{\$4.28} \\
\midrule
Active Exploration Baseline
    & 47.2 & 25.8K & 3.5K & 29.3K & \$9.40 \\
SAVE
    & \underline{57.1} & 28.5K & \underline{1.7K} & 30.3K & \$7.80 \\

\midrule
\midrule
\midrule

\multicolumn{6}{l}{\textbf{Gemma 4}} \\

\addlinespace[2pt]
\multicolumn{6}{l}{\emph{Minerva}} \\
Gemma 4 Native~\citet{gemma4}
    & 51.6 & \underline{15.6K} & \underline{0.3K} & \underline{15.9K} & \\
VideoTree~\citet{wang2025videotree}
    & 51.6 & 33.0K & 3.4K & 36.4K & \\
VideoAgent~\citet{VideoAgent}
    & 42.3 & 21.2K & 1.1K & 22.3K & \\
VideoLucy~\citet{zuo2025videolucy}
    & \underline{58.4} & 65.4K & 6.6K & 72.0K & \\
VideoCuriousAgent~\citet{yang2025vca}
    & 36.4 & 25.1K & 1.3K & 26.4K & \\
DeepVideoDiscovery~\citet{zhang2025deep}
& 52.5 & 58.4K & 13.7K & 72.0K & \\
A.I.R.~\citet{zou2026air}
    & 51.9 & 20.0K & 2.1K & 22.1K & \\    
SAOS
& 52.5 & \textbf{8.6K} & \textbf{0.3K} & \textbf{8.9K} & \\
\midrule
Active Exploration Baseline
    & 52.6 & 26.8K & 3.6K & 30.4K \\
SAVE
    & \textbf{62.4} & 18.9K & 0.9K & 19.7K & \\

\midrule
\midrule

\addlinespace[4pt]
\multicolumn{6}{l}{\emph{EgoSchema}} \\
Gemma 4 Native~\citet{gemma4}
    & 72.4 & 15.7K & \textbf{0.2K} & 16.0K & \\
VideoTree~\citet{wang2025videotree}
    & 71.1 & 31.4K & 3.0K & 34.4K & \\
VideoAgent~\citet{VideoAgent}
    & 67.7 & \textbf{4.7K} & 0.4K & \textbf{5.1K} & \\
VideoLucy~\citet{zuo2025videolucy}
    & \underline{73.2} & 89.5K & 14.2K & 103.6K & \\
VideoCuriousAgent~\citet{yang2025vca}
    & 67.7 & 9.3K & 0.6K & 9.9K & \\
DeepVideoDiscovery~\citet{zhang2025deep}
    & 71.5 & 109.9K & 12.3K & 122.2K & \\
A.I.R.~\citet{zou2026air}
    & \textbf{74.3} & 21.1K & 1.9K & 23.0K & \\
SAOS
    & 70.5 & 9.7K & \underline{0.2K} & 9.9K & \\
\midrule
Active Exploration Baseline
    & 65.3 & 9.4K & 1.2K & 10.6K \\
SAVE
    & 72.4 & \underline{9.0K} & {0.4K} & \underline{9.4K} & \\

\midrule
\midrule

\addlinespace[4pt]
\multicolumn{6}{l}{\emph{VSIBench (tiny)}} \\
Gemma 4 Native~\citet{gemma4}
    & \underline{49.1} & 15.6K & \underline{0.3K} & 15.9K & \\
VideoTree~\citet{wang2025videotree}
    & 42.8 & 22.3K & 2.4K & 24.6K & \\
VideoAgent~\citet{VideoAgent}
    & 44.5 & 25.1K & 1.1K & 26.2K & \\
VideoLucy~\citet{zuo2025videolucy}
    & 43.8 & 59.3K & 4.1K & 63.4K & \\
VideoCuriousAgent~\citet{yang2025vca}
    & 41.4 & 22.9K & 1.3K & 24.1K & \\
DeepVideoDiscovery~\citet{zhang2025deep}
    & 49.1 & 29.8K & 4.8K & 34.6K & \\
A.I.R.~\citet{zou2026air}
    & 45.7 & 18.3K & 1.9K & 20.2K & \\
SAOS
& 43.5 & \textbf{8.3K} & \textbf{0.3K} & \textbf{8.6K} & \\
\midrule
Active Exploration Baseline
    & 45.2 & 27.2K & 3.8K & 31.0K \\
SAVE
    & \textbf{52.6} & \underline{10.5K} & 0.6K & \underline{11.1K} & \\

\bottomrule
\end{tabular}
\end{table*}

\section{SAVE Prompts and Algorithm}
\label{sec:prompts}

\subsection{Algorithm}
\begin{algorithm}[H]
\caption{SAVE: Simple Agentic Video Explorer}
\label{alg:save}
\begin{algorithmic}[1]
\Require Video $V={x_i}_{i=1}^{N}$, question $q$, backbone $M$,
query budget $B$, maximum returned scenes $K$,
compression interval $C$, high-resolution threshold $\delta$

\State $h \gets [\text{video metadata},q]$
\Comment{Text-only exploration history}
\State $(r_0,a_1) \gets M(h)$
\Comment{Initial reasoning and action}
\State $h \gets \Call{Append}{h,r_0}$

\For{$t=1,\ldots,B$}
\If{$a_t=\textsc{Stop}$}
\State \textbf{break}
\EndIf

\State $(s,e) \gets a_t$
\Comment{$a_t=\textsc{QueryRange}(s,e)$}

\State $\mathcal{S} \gets
    \Call{SceneDetect}{V[s:e]}$
\Comment{Adapt the detection threshold to the queried range}

\State $\mathcal{S}' \gets
    \Call{UniformSubsample}{\mathcal{S},K}$
\Comment{Restrict number of scenes to K}
\State $I \gets
    \{\operatorname{mid}(S)\mid S\in\mathcal{S}'\}$

\State $d_t \gets |e-s|/\operatorname{fps}(V)$
\State $\rho_t \gets
    \begin{cases}
        \textsc{High}, & d_t \leq \delta,\\
        \textsc{Medium}, & d_t > \delta
    \end{cases}$
\Comment{Use higher visual fidelity for narrow ranges}

\State $F_t \gets \Call{LoadFrames}{V,I,\rho_t}$
\Comment{Load selected frames and transcripts at resolution $\rho_t$}
\State $o_t \gets
    \{(i,F_t[i])\mid i\in I\}$

\State $(r_t,a_{t+1}) \gets M(q,h,o_t)$
\Comment{Observe, reason, and select the next query range}
\State $h \gets \Call{Append}{h,r_t}$

\If{$t \bmod C=0$}
    \State $h \gets \Call{Compress}{h,q}$
    \Comment{Retain relevant evidence and searched-range anchors}
\EndIf

\EndFor

\State $\hat{y} \gets M.\Call{FinalAnswer}{q,h}$
\State \Return $\hat{y}$
\end{algorithmic}
\end{algorithm}

\subsection{Prompts}

\textbf{Agent prompt.}
The active vision agent was prompted to iteratively gather visual evidence before answering.
At each step, it received the video metadata, the question, and the following action interface:

\begin{quote}
You are an Active Vision Agent.

Answer the video question using as few visual queries as needed.
Cost matters, but accuracy matters more; gather enough evidence before answering.

Video: [number of frames] frames, FPS [fps], duration [duration]s.
Frame indices are original frame numbers from 0 to [last frame index];
timestamp = index / [fps].

Output exactly: \textbf{Observation}, \textbf{Thought}, then one \textbf{Action}.

\textbf{Observation:} Describe only relevant facts from the new frames, with exact frame indices.

\textbf{Thought:} Explain the next most useful evidence to gather or why you can answer.

\textbf{Action:} choose exactly one:
\begin{itemize}
    \item \texttt{ACTION: QUERY <start\_index> <end\_index>}: inspect that frame range and return relevant visual observations plus transcript. Example: \texttt{ACTION: QUERY 0 3000}.

    A QUERY returns selected sampled frames from the requested range, not every frame. For broader queried ranges, fine details or brief events may be missed between sampled frames. If answer-changing details remain uncertain, query a narrower range.

    \item \texttt{ACTION: STOP}: stop only when you're certain. Do not STOP if two answers remain plausible from the visual evidence. If any answer-changing uncertainty remains, use \texttt{QUERY}.
\end{itemize}

Sometimes a precise answer might not be possible (e.g., numerical questions); in that case, always provide your best estimate and stop once you're reasonably confident. Never provide an empty or invalid answer.

Question: ``[question]''

Do not hallucinate.
\end{quote}

\paragraph{History compression prompt.}
Interaction history was compressed with a separate prompt that preserved navigation anchors while removing irrelevant visual detail:

\begin{quote}
You are an expert at information compression. Compress the interaction history below. The agent is answering the given question.

\textbf{Rules:}
\begin{enumerate}
    \item Preserve every \texttt{[OBSERVATION: Indices ...]} header exactly; these are navigation anchors showing which video segments were searched.
    \item For observations where the agent found nothing relevant, replace the body with one short phrase, e.g. ``Irrelevant --- <3-word reason>.'' No visual details.
    \item For observations where the agent found something relevant, write 1--2 terse sentences capturing only the key facts that help answer the question. Drop all descriptions irrelevant to the question. Keep precise frame indices when relevant.
    \item End with a single \texttt{Status:} line summarising the current hypothesis and what region to search next.
    \item Output only the compressed result, with no preamble or meta-commentary.
\end{enumerate}
\end{quote}

\section{Further Analysis}

\subsection{Statistical Significance}
\label{sec:statistical_significance}

On pairwise common error-free Minerva samples, two-sided video-cluster permutation tests found that SAVE outperformed VideoLucy, VideoTree, VideoCuriousAgent, VideoAgent, DeepVideoDiscovery, A.I.R., and SAOS. All seven comparisons remain significant after Holm correction across all eight baselines (\(p_{\mathrm{Holm}}\leq 8.0\times10^{-5}\)); the 100,000-permutation run could not resolve smaller p-values. The difference between SAVE and direct Gemini 3 Flash is not statistically significant (\(p_{\mathrm{Holm}}=0.390\)).

On the 491 common error-free EgoSchema samples, SAVE achieved the highest accuracy of 75.97\%, significantly outperforming VideoAgent (64.2\%; \(p_{\mathrm{Holm}} = 3.25 \times 10^{-6}\)) and DeepVideoDiscovery
(69.0\%; \(p_{\mathrm{Holm}} = 0.0052\)) under paired, two-sided exact McNemar tests. Its improvements over VideoLucy (72.7\%; \(p_{\mathrm{Holm}} = 0.4354\)), VideoTree (73.1\%; \(p_{\mathrm{Holm}} = 0.4354\)), A.I.R. (73.9\%; \(p_{\mathrm{Holm}} = 0.736\)) and
VideoCuriousAgent (72.3\%; \(p_{\mathrm{Holm}} = 0.177\)) were not statistically significant, however SAVE required
approximately 46--83\% fewer tokens than these methods. SAVE  also performed comparably to both non-iterative baselines, direct Gemini~3 Flash
inference (76.0\% versus 75.4\%; \(p_{\mathrm{Holm}} = 0.826\)) and SAOS (73.1\%; \(p_{\mathrm{Holm}} = 0.4354\)), although both methods were more efficient, requiring 13.7K and 19.3K versus 22.5K tokens per sample respectively.

For the Gemma~4 evaluation on Minerva, SAVE achieved the highest accuracy (62.4\%). In paired, two-sided permutation tests clustered by video using 100,000 permutations, SAVE outperformed all eight baselines after Holm correction. The adjusted p-value was \(8.0\times10^{-5}\) for A.I.R., VideoTree, VideoCuriousAgent, VideoAgent, direct Gemma~4 inference, DeepVideoDiscovery, and SAOS. This is the smallest Holm-adjusted value resolvable with 100,000 permutations across eight comparisons. For VideoLucy, $p_{\mathrm{Holm}} = 0.0345$. SAVE used 19.7K tokens per question versus 72.0K for VideoLucy, a 72.6\% reduction.

For Gemma on EgoSchema, SAVE achieved 72.4\% accuracy, with no statistically significant difference from any of the eight baselines after Holm correction. Under paired, two-sided exact McNemar tests, SAVE performed comparably to VideoLucy ($p_{\mathrm{Holm}} = 1.000$), direct Gemma~4 inference ($p_{\mathrm{Holm}} = 1.000$), DeepVideoDiscovery ($p_{\mathrm{Holm}} = 1.000$), VideoTree ($p_{\mathrm{Holm}} = 1.000$), SAOS ($p_{\mathrm{Holm}} = 1.000$), A.I.R. ($p_{\mathrm{Holm}} = 1.000$), VideoAgent ($p_{\mathrm{Holm}} = 0.168$), and VideoCuriousAgent ($p_{\mathrm{Holm}} = 0.139$). Despite the statistically comparable performance, SAVE remained substantially more token-efficient than most prior methods, requiring only 9.4K tokens per sample compared with 34.4K for VideoTree, 103.6K for VideoLucy, 9.9K for VideoCuriousAgent, 16.0K for direct Gemma~4 inference, and 122.0K for DeepVideoDiscovery.

\section{Synthetic evidence-locality benchmark}
\label{sec:synthetic_benchmark}

Our benchmark contains 384 procedurally generated videos (21.6 hours in total), rendered at $256\times256$ resolution and 30 frames per second. Each video shows a stationary square against a white background, repeatedly changing among red, green, blue, and yellow in a fixed order. We define one complete four-color cycle as an \emph{event}: observing a full cycle is sufficient to recover the order. Cycle duration is varied across 2, 5, 10, and 15 seconds, with each color displayed for approximately one quarter of the cycle. Video duration is varied independently across 30, 60, 120, and 600 seconds. For each video-duration and cycle-duration pair, we generate 24 videos: four instances of each of the six possible color orders anchored at red. The clip starts at a randomly selected point in the cycle, so its first visible color need not be red. Each sample asks for the order starting with red and presents five shuffled choices: the correct order and four alternatives drawn from the remaining five. Because the same event repeats throughout the video, locating the relevant event is trivial. This allows us to study the benefit of adaptive inspection independently of temporal localization. We will publicly release the videos, annotations, and generation code to support further research.

\begin{figure}[htbp]
    \centering
    \fbox{\includegraphics[width=\textwidth]{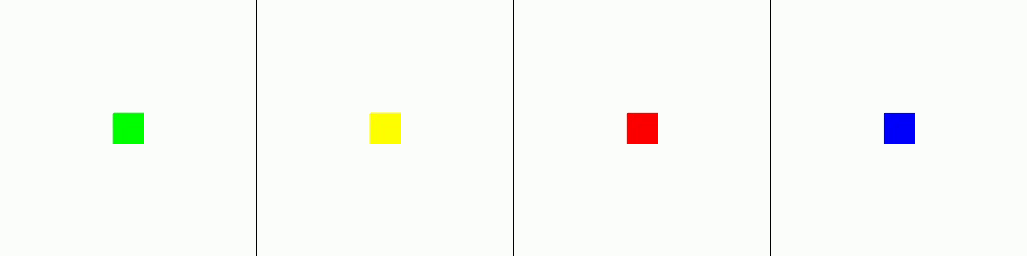}}
    \caption{Illustration of our synthetic benchmark using video 3. From left to right, frames are sampled at 0, 6, 12 and 18s.}
    \label{fig:synthetic_dataset_illustration}
\end{figure}

\clearpage
\section{Case Studies}
\label{sec:case_studies}

\begin{figure}[htbp]
    \centering
    \includegraphics[width=\textwidth]{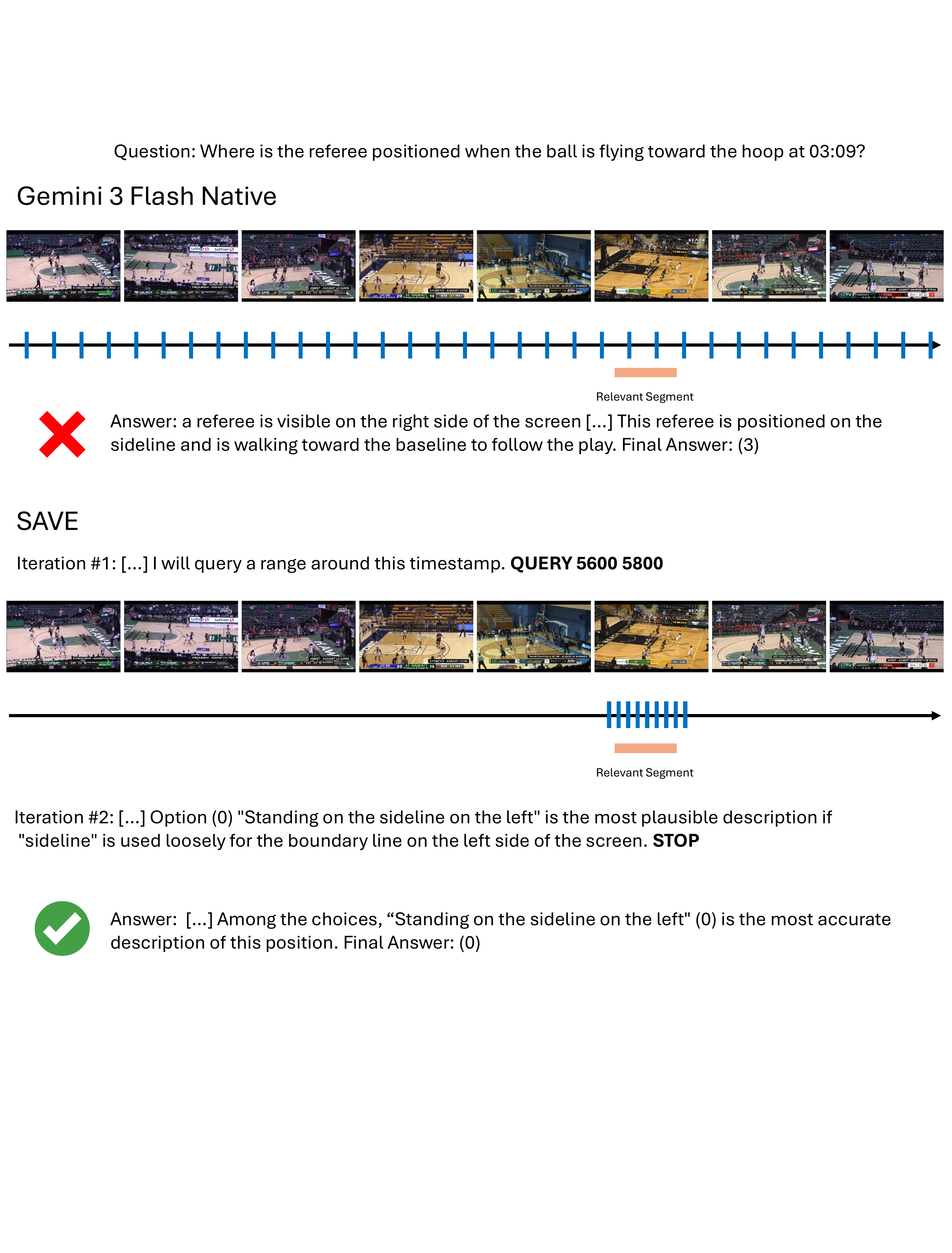}
    \caption{Example of both Gemini 3 Flash Native and SAVE answering a question from Minerva (question \#7). In this example a specific timestamp is directly given in the question. Gemini simply downsamples the whole video to 1 FPS and therefore has no possibility of efficiently querying only the relevant segment. SAVE uses its range query to specifically observe the relevant part at high granularity and correctly answers the question.}
    \label{fig:case_study_1}
\end{figure}

\begin{figure}[htbp]
    \centering
    \includegraphics[width=\textwidth]{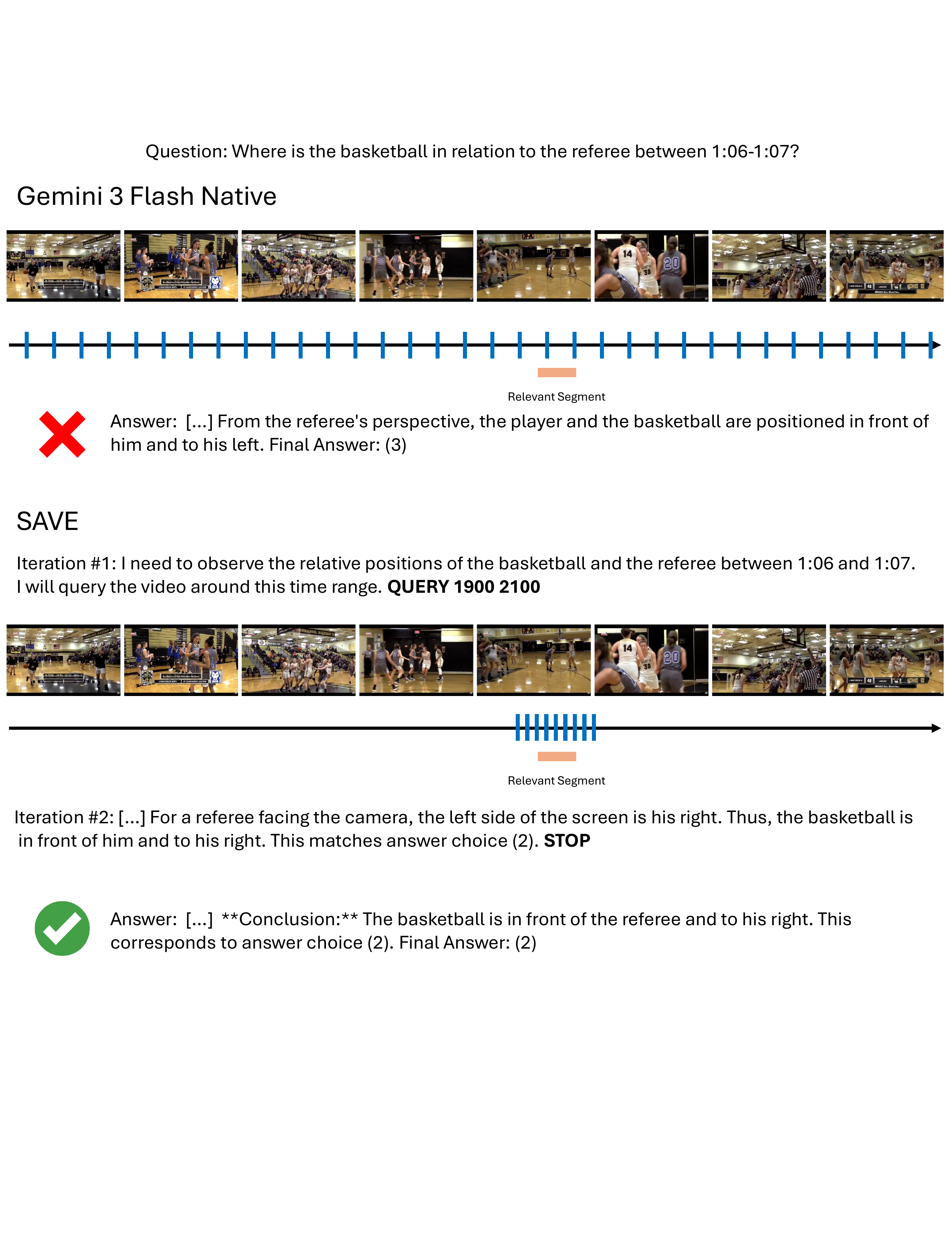}
    \caption{Example of both Gemini 3 Flash Native and SAVE answering a question from Minerva (question \#24). In this example a specific timestamp is directly given in the question. Gemini simply downsamples the whole video to 1 FPS and therefore has no possibility of efficiently querying only the relevant segment. SAVE uses its range query to specifically observe the relevant part at high granularity and correctly answers the question.}
    \label{fig:case_study_2}
\end{figure}

\begin{figure}[htbp]
    \centering
    \includegraphics[width=\textwidth]{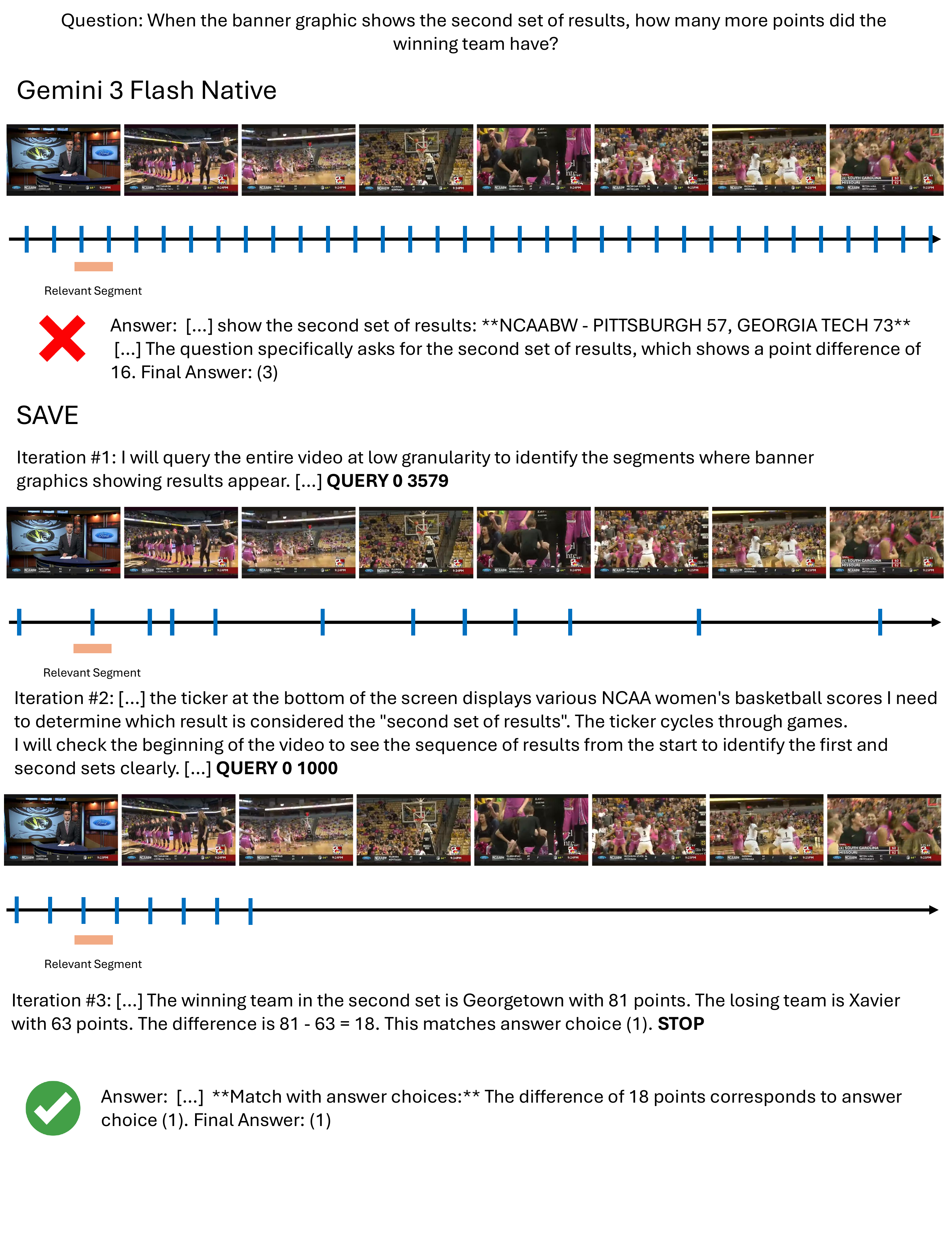}
    \caption{Example of both Gemini 3 Flash Native and SAVE answering a question from Minerva (question \#34). No timestamp is given in the question, so both methods first have to find the relevant moment in the video. SAVE employs a multi-step search strategy: First, it queries the whole video at low granularity, subsequently isolates the crucial segment from the rest of the video and answers the question correctly. Gemini's native video processing instead passes the whole video into the model at a higher granularity than SAVE, but nonetheless fails to answer the question correctly, as fine, crucial details get lost in the large amount of irrelevant context passed into the model.}
    \label{fig:case_study_3}
\end{figure}

\begin{figure}[htbp]
    \centering
    \includegraphics[width=\textwidth]{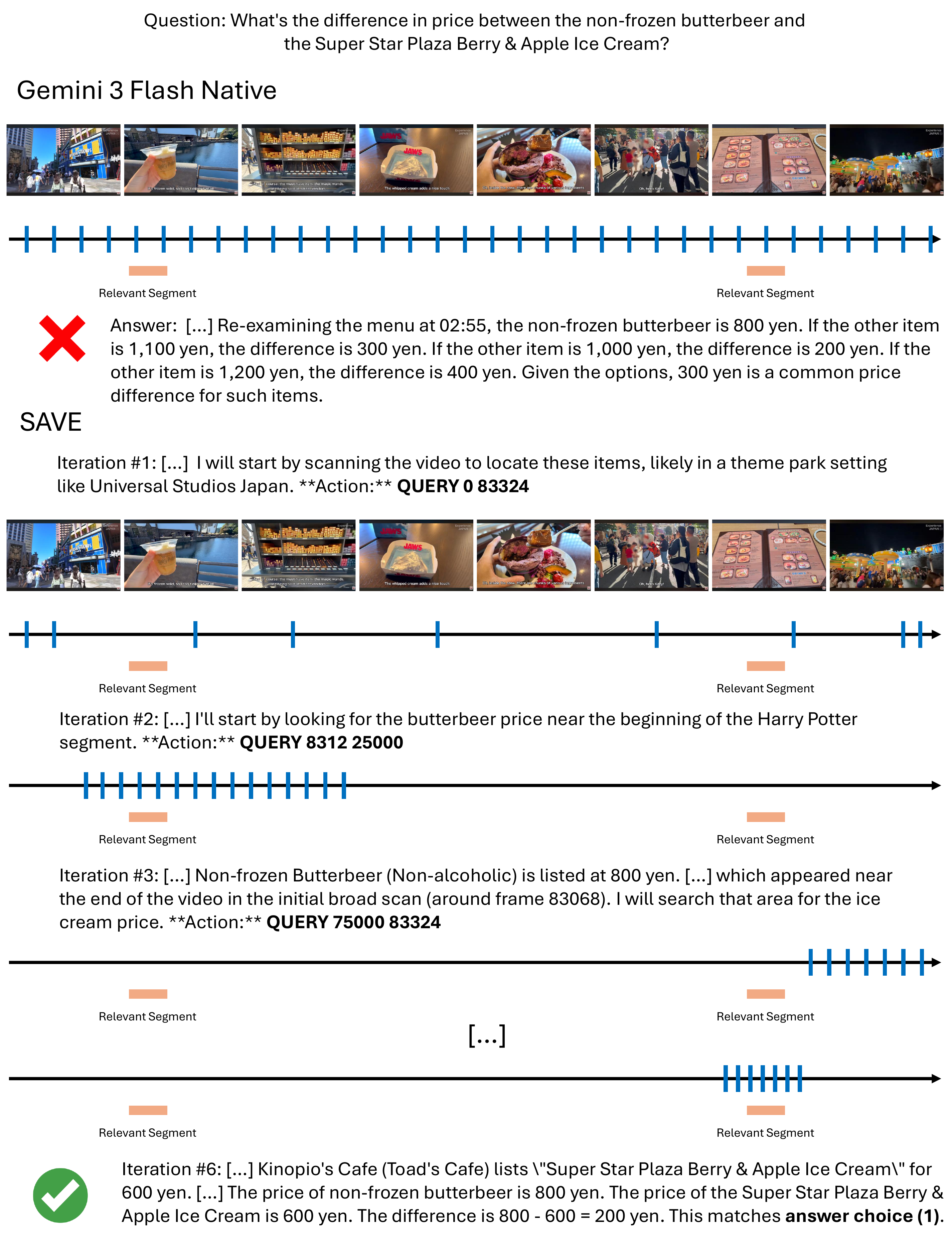}
    \caption{Example of both Gemini 3 Flash Native and SAVE answering a question from Minerva (question \#1357). In this case, two moments from the video are crucial to answering the question. Neither timestamp is given in the question. SAVE employs a multi-step search strategy: First, it queries the whole video at low granularity and subsequently isolates the first crucial segment from the rest of the video. After having correctly interpreted the first relevant segment, it proceeds to search for the second relevant segment later in the video. After 6 iterations, the second relevant moment is found and SAVE answers the question correctly. Gemini's native video processing instead passes the whole video into the model, manages to correctly identify the first moment, but the second moment is lost among the large amount of irrelevant visual context presented to the model.}
    \label{fig:case_study_4}
\end{figure}

\clearpage

\end{document}